\documentclass{article}
\PassOptionsToPackage{numbers,sort&compress}{natbib}
\usepackage[preprint]{neurips_2026}

\usepackage[utf8]{inputenc}
\usepackage[T1]{fontenc}
\usepackage{amsmath,amssymb,amsfonts,amsthm}
\usepackage{graphicx}
\usepackage[table]{xcolor}
\usepackage{booktabs,multirow,makecell,array}
\usepackage{algorithm,algorithmic}
\usepackage{threeparttable,placeins}
\usepackage{float}
\usepackage{microtype}
\usepackage{url}
\usepackage[hypertexnames=false]{hyperref}

\newtheorem{theorem}{Theorem}

\colorlet{proursbg}{black!6}

\title{Structured Confidence Perturbation via Fisher Geometry for Adversarial Training}
\author{%
  Chenlong Wei \\
  Xi'an Jiaotong-Liverpool University \\
  China \\
  \texttt{Chenlong.Wei23@student.xjtlu.edu.cn}
  \And
  Hanling Wang \\
  University of California, San Diego \\
  United States \\
  \texttt{haw168@ucsd.edu}
  \AND
  Chong Zhang \\
  Xi'an Jiaotong-Liverpool University \\
  China \\
  \texttt{Chong.Zhang19@student.xjtlu.edu.cn}
  \And
  Qinkai Yu \\
  University of Exeter \\
  England \\
  \texttt{qy269@exeter.ac.uk}
  \AND
  Fangyu Wu \\
  Xi'an Jiaotong-Liverpool University \\
  China \\
  \texttt{Fangyu.Wu02@xjtlu.edu.cn}
  \And
  Xiaobo Jin\thanks{Corresponding author.} \\
  Xi'an Jiaotong-Liverpool University \\
  China \\
  \texttt{Xiaobo.Jin@xjtlu.edu.cn}
}

\begin{document}
\maketitle

\begin{abstract}
Adversarial training is often plagued by ``pseudo-attacks''---perturbations that significantly increase the loss value but fail to truly alter the model's confidence structure. To address this, we propose two complementary approaches to directly attack and defend against confidence structure deformation. First, building upon the traditional loss maximization perspective, we introduce a hybrid objective that integrates cross-entropy and KL divergence, encouraging adversarial examples to simultaneously disrupt prediction accuracy and distribution consistency. Second, we geometrically reconstruct adversarial training by generating perturbations that maximize the distribution distance on the confidence manifold. Theoretical analysis shows that the optimal perturbation direction aligns with the principal eigenvectors of the Fisher information matrix. We also propose an efficient algorithm to avoid high-dimensional matrix computations. Experiments on multiple datasets and model architectures demonstrate that our method outperforms existing methods in adversarial robustness and robust transferability, particularly excelling in structural perturbations within high-confidence regions. Code repository: \url{https://github.com/wnzgy/SCP}.
\end{abstract}

\section{Introduction}
\label{sec:intro}
Deep neural networks (DNNs) have achieved remarkable progress across numerous applications; however, despite their impressive performance, they remain highly susceptible to adversarial perturbations which conduct imperceptible modifications to inputs that cause incorrect predictions \cite{chakraborty2021survey}. This vulnerability has raised significant concerns in safety--critical domains such as image recognition \cite{huang2017adversarial}, autonomous driving \cite{kabid2023autonomous}, and anomaly detection \cite{zhu2022adversarial}. Among various defense strategies, \textit{adversarial training} has emerged as one of the most effective methods for improving model robustness \cite{kraidia2024defense,zhou2022adversarial}. By jointly optimizing over clean and adversarial samples, it encourages models to learn stable decision boundaries under input perturbations.

% % 现有方法介绍和不足
Existing adversarial training methods are commonly grouped into gradient-based approaches, optimization-based formulations, and attack-driven adversarial sample generation. Gradient-based methods such as FGSM and PGD \cite{goodfellow2014explaining,madry2019deeplearningmodelsresistant} construct perturbations by following the loss gradient, offering high computational efficiency but ensuring robustness only in a local linear neighborhood.  Optimization-based defenses treat adversarial training as a min--max problem and explicitly minimize the worst-case loss under bounded perturbations, often using solvers like L-BFGS \cite{liu1989limited} or zero-order optimization \cite{duchi2015optimal}; however, their computational overhead is typically prohibitive for large-scale models. Meanwhile,  adversarial attack--based methods \cite{li2022review} generate perturbed samples directly by designing black-box or white-box attacks: black-box methods rely solely on model outputs, whereas white-box attacks exploit full model gradients and internal structures.

% %不足
% Although such attacks effectively expose model vulnerabilities, defending against them remains challenging due to their rapidly evolving and diverse perturbation strategies. Despite their differences, these three families of methods all share a fundamental paradigm: adversarial examples are generated by maximizing the classification loss of the true label. Yet, loss maximization mainly suppresses the discriminant value of the ground-truth class and often fails to induce meaningful, global changes in the entire confidence distribution. As illustrated in Fig.~\ref{fig:abs}, even when the true label's logit is reduced, its relative confidence may remain higher than that of other classes---a phenomenon we refer to as the pseudo-attack. This reveals a core limitation of existing perturbation approaches: maximizing the loss through a pseudo-attack does not necessarily maximize the structural shift in the model's output distribution.

\begin{figure}[htbp]
    \centering
    \includegraphics[width=1.0\linewidth]{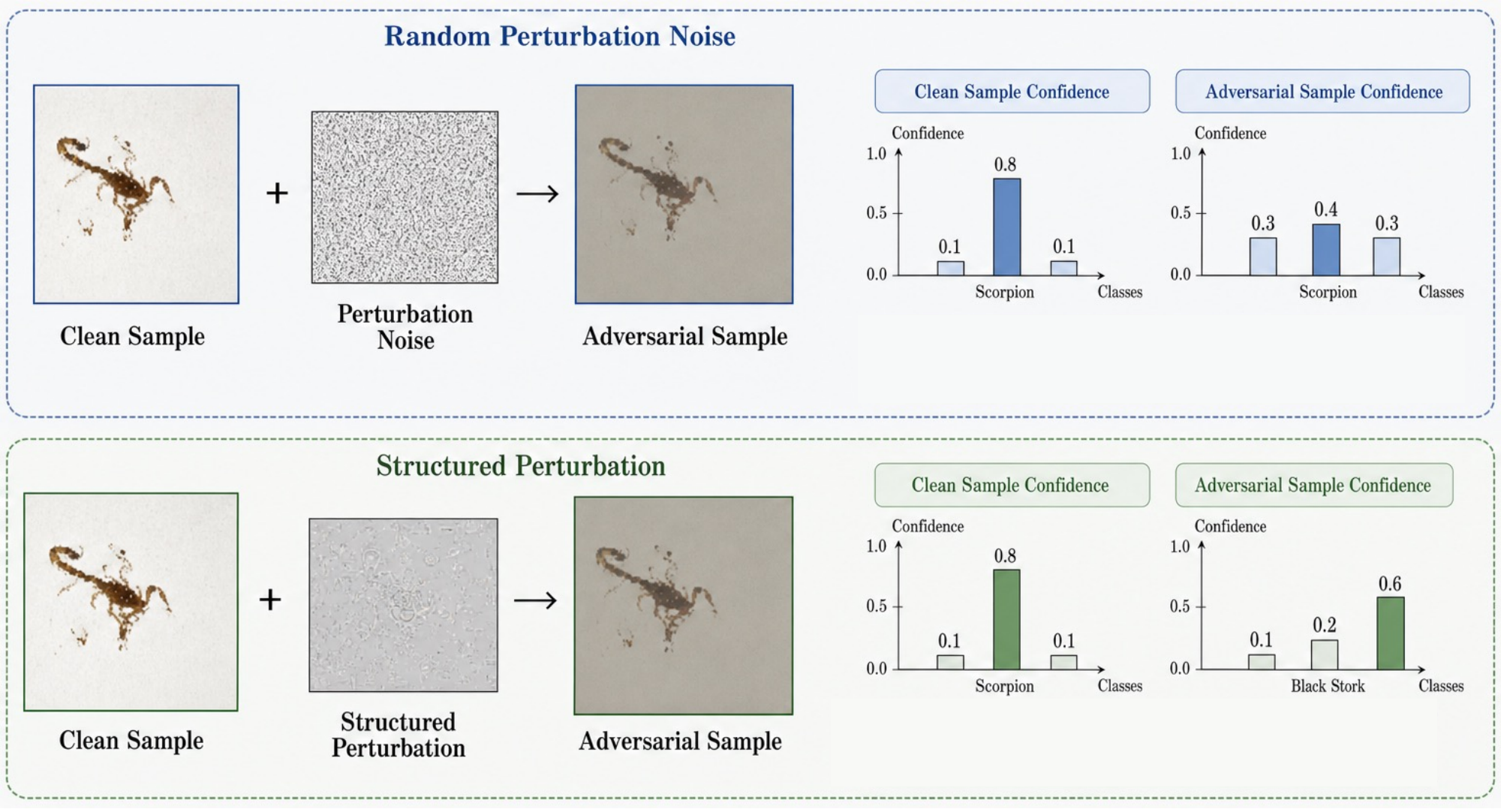}
    \caption{Most previous methods add perturbation noise to a single discriminant value $f(x)$ for adversarial training (top), while our SCP adds structured perturbations to the statistical manifold $\{p(y|x;\theta)|x \in X\}$ for adversarial training (bottom).}
    \label{fig:abs}
    % \vspace{-15pt}
\end{figure}

Prevailing adversarial training methods largely share a common heuristic: they craft perturbations by maximizing the classification loss with respect to the true label. While effective in reducing the predicted probability of the true class, this loss-centric paradigm often fails to induce meaningful, global changes in the model's output distribution. As illustrated in Fig.~\ref{fig:abs}, even when the loss is increased, the relative ordering of confidence across classes may remain intact---a phenomenon we term pseudo-attacks. We define a "pseudo-attack" as a perturbation $\delta$ satisfying $\mathcal{L}(x+\delta,y) > \mathcal{L}(x,y)$ but $\arg\max_k p(y=k|x+\delta) = \arg\max_k p(y=k|x)$. Even if the top-1 remains unchanged, the relative confidence ranking and distribution shape can still maintain local stability, meaning that the geometric vulnerability of the model near the decision boundary is not truly challenged. Such perturbations do not genuinely challenge the structural geometry of the model's confidence manifold, leaving underlying vulnerabilities exposed.

To move beyond local loss maximization, we argue that adversarial robustness should be grounded in the geometric stability of the model's confidence distribution. In this paper, we introduce a unified adversarial training framework that operates directly on the statistical manifold of output probabilities. 
\noindent Our main contributions are summarized as follows:
\begin{itemize}
\item[$\bullet$] We propose an enhanced Single Point Perturbation (SPP) strategy that combines cross-entropy (CE) and Kullback--Leibler (KL) divergence into a hybrid objective. Unlike prior uses of KL as a regularizer \cite{zhang2019theoretically}, we deploy it actively during attack generation to disrupt both the top-1 prediction and the overall confidence structure.

\item[$\bullet$] We develop a SCP framework that reformulates adversarial search as maximizing the distance between probability distributions on the confidence manifold. We show theoretically that the optimal perturbation aligns with the principal eigenvector of the Fisher Information Matrix (FIM), which captures the local curvature of the manifold. To enable scalable computation, we derive an efficient eigen-decomposition that avoids explicit manipulation of high-dimensional FIM.

\item[$\bullet$] Extensive experiments on CIFAR-10, CIFAR-100, and Tiny-ImageNet demonstrate that our methods consistently improve adversarial robustness and enhance attack transferability across architectures. By directly targeting the geometric roots of model vulnerability, SCP and SPP advance the state of the art in adversarial training, offering a principled and computationally feasible path toward truly robust deep learning systems.
\end{itemize}

\section{Related work}
\label{sec:related}

Adversarial Training and Geometry.  Adversarial training (AT) has evolved from a local noise defense strategy into a holistic paradigm that reshapes the decision manifold of deep neural networks. Early mainstream gradient-based methods, such as FGSM \cite{goodfellow2014explaining} and PGD \cite{madry2019deeplearningmodelsresistant}, primarily constructed adversarial perturbations within a local $\epsilon$-sphere along the direction of steepest loss increase. While effective, their robustness was often limited to these direct input neighborhoods. To seek more globally stable solutions, optimization-based frameworks \cite{madry2019deeplearningmodelsresistant,zhang2019theoretically} modeled adversarial training as a minimax game. Although subsequent acceleration strategies, such as FreeAT \cite{shafahi2019adversarial}, YOPO \cite{zhang2019you}, improved training efficiency, they essentially still followed the same loss-centric heuristic. In recent years, some methods, including feature scattering \cite{zhang2019defense} and MART \cite{wang2020improving}, have attempted to learn more transferable features through feature space alignment or hard example reweighting, but they often neglect the underlying geometry of the data manifold. 

Many existing approaches rely on maximizing the cross-entropy (CE) loss to measure adversarial vulnerability, while methods such as TRADES maximize distributional discrepancy during attack generation. This approach is essentially a ``pointwise suppression'' of the true logit, which typically fails to induce a global distortion of the model's confidence structure. While TRADES \cite{zhang2019theoretically} and ensemble-based approaches \cite{tramer2017ensemble} introduce KL divergence or diversity constraints, they generally treat distributional disparity as a scalar penalty rather than a structural measure. Our work breaks with this convention by redefining adversarial perturbations as geometric transformations on the confidence manifold and asserting that true robustness stems from the manifold's geometric stability under structural stress.

Information-Theoretic Perspectives. The application of KL divergence in adversarial training has traditionally been limited to its use as a scalar measure of distributional disparity. A common practice is to use it as a consistency regularization term to balance the tradeoff between natural accuracy and adversarial robustness (e.g., TRADES \cite{zhang2019theoretically} or gradient-smoothing approaches \cite{andriushchenko2020understanding}). Another perspective stems from knowledge distillation, where KL divergence guides distribution alignment  \cite{wu2025rethinking}, thereby facilitating the learning of robust representations. However, most of these paradigms overlook the profound informational geometric meaning of KL divergence. Within the informational geometry framework, the KL divergence between two neighboring probability distributions is equivalent to a second-order approximation of the Mahalanobis distance defined by the Fisher information matrix (FIM).

Existing defense methods utilize its size for constraint but fail to fully leverage its potential as a tool for discovering structured vulnerabilities. Our research bridges this gap by transforming KL divergence from a passive regularization term into an active target for attack generation. By maximizing the manifold distance along the Fisher principal eigenvectors, we can systematically disrupt the global structure of the confidence distribution, rather than merely suppressing local outputs, forcing the model to learn a more resilient geometric representation and effectively overcoming the evaluation blind spots inherent in traditional paradigms.

\section{Methodology}

Traditional adversarial training generates perturbations by maximizing the loss function (e.g., cross-entropy CE), but such methods only locally suppress the logit of the true class, ignoring the global geometry of the confidence distribution (Fig. \ref{fig:abs}). We propose that true robustness should stem from the model's stability on the confidence manifold, i.e., the input perturbation should maximize the structural deformation of the output probability distribution. To this end, we design two complementary strategies:

\begin{itemize}
\item[$\bullet$] Local Single-Point Perturbation (SPP): Jointly optimizes CE and KL divergence, simultaneously disrupting prediction accuracy and distribution consistency;

\item[$\bullet$] Global Structured Confidence Perturbation (SCP): From an information geometry perspective, it generates perturbations using the principal eigendirections of the Fisher Information Matrix (FIM), directly manipulating the curvature of the confidence manifold. 
\end{itemize}

\subsection{Geometric motivation: from single points to manifold deformation}

Let the model be $f_{\theta}: \mathcal{X} \rightarrow R^{K}$, where $\theta$ denotes the parameters and $K$ is the number of classes. The softmax output $p(y|x;\theta)$  defines a statistical manifold. Each input $x$ corresponds to a point on $\mathbb{P}$, and a perturbation $\delta$ moves $x$ to $x' = x + \delta$, inducing a trajectory on the manifold. Traditional adversarial training aims to maximize the cross-entropy loss $\mathcal{L}_{\text{CE}}(y, x')$, which only affects the probability of the true class. In contrast, we propose to maximize a distance metric $\text{KL}(p(y|x),p(y|x'))$ on $\mathbb{P}$, which captures global changes in the confidence structure.

\subsection{Adversarial training with single point perturbation (SPP)}

% Previous work TRADES \cite{zhang2019theoretically} proposed a prominent defense method and the objective function can be defined as follows:

Previous TRADES \cite{zhang2019theoretically} introduces a defense objective that balances accuracy and robustness via a KL-divergence regularizer

\begin{equation}
\arg \min_{\theta} [  \mathcal{L}_{\text{CE}}(y, f_{\theta}(x)) + \lambda \max_{\delta : \|\delta\|_p \leq \epsilon} \text{KL}(f_{\theta}(x), f_{\theta}(x + \delta)) ],
\end{equation}
where CE is the cross-entropy loss, KL is the Kullback-Leibler divergence and $\lambda$ controls the trade-off between accuracy and robustness. 

% It is worth noting that the maximization of the cross-entropy loss in the K-class classification problem can be represented as the maximization problem of the KL divergence
% \vspace{-10pt}
% \begin{align}
%     \small
%     \max_{x' = x + \delta} L(x',y; \theta) & = \max_{x' = x + \delta} - \sum_{i = 1}^K y_i \log p(y_i|x';\theta) \nonumber \\
%     & = \mathrm{KL}(\delta(y = 1|x),p(y|x';\theta)),
% \end{align}
% where the groundtruth distribution of the label $\delta(y=1|x)$ is defined as $\delta(y_i = 1|x) = 1, i = 1,2,\cdots,K.$

Unlike the TRADES~\cite{zhang2019theoretically} method, we repurpose this formulation for attack generation. Specifically, we define a hybrid loss (as shown in Alg. \ref{alg:spp})

\begin{equation}
    \mathcal{L}_{\text{SPP}}(x',x,y) = \mathcal{L}_{\text{CE}}(y,f_{\theta}(x')) + \text{KL}(f_{\theta}(x),f_{\theta}(x')),
\end{equation}
and generate adversarial examples by maximizing $\mathcal{L}_{\text{SPP}}$ within an $\epsilon-$ball
\begin{equation}
    x' = x + \arg\max_{\|\delta\| \le \alpha} \mathcal{L}_{\text{SPP}}(x + \delta,x,y),
\end{equation}
where the first term encourages misclassification, while the second term disrupts the entire output distribution, preventing pseudo-attacks that leave the confidence ordering intact.

% which uses them for defense, we combine CE and KL to generate adversarial examples. This means we want the adversarial examples to not only misclassify the model but also disrupt the structure of the output distribution (see Alg. \ref{alg:spp}).

\begin{algorithm}
\caption{Adversarial training with single point perturbation (SPP)}
\label{alg:spp}
% \footnotesize
% \small
\begin{algorithmic}[1]
   \STATE Input: DNN model $f_\theta(x)$; dataset $\mathcal{D}$ with $K$ classes; batch size $m$; learning rate $\eta$; radius for adversaries $\epsilon$;  step size $\alpha$ and $S$ for attack iterations; number of epochs $T$.
   \STATE Randomly initialize the network parameter $\theta$ and $\delta \sim 0.001 \cdot \mathcal{N}(0, 1)$
   \FOR {$t = 0,1,2,\cdots,T$}
     \FOR {each batch $\mathcal{B} = \{(x_i, y_i)\}_{i=1}^m$ in $\mathcal{D}$}
        \STATE  $\tilde{\mathcal{B}} = \emptyset$
       \FOR{$(x_i, y_i)$ in $\mathcal{B}$}
        \STATE $\hat x_i$ = $x_i + \delta$
        \FOR{$s = 1, 2, \dots, S$}
            \STATE$\mathcal{L}_{\text{SPP}} = \mathcal{L}_{\text{CE}}\left( y_i, f_{\theta^{}} (\hat{x}_{i}) \right)+ \, \text{KL} \left( f_{\theta^{}} (x_{i}), f_{\theta^{}} (\hat{x}_{i}) \right)$
            \STATE $\hat{x}_i \leftarrow  \hat{x}_{i} + \alpha \cdot \text{sign}  \left(\nabla_{\hat{x}_{i}}\mathcal{L}_{\text{SPP}}  \right)$ 
        \ENDFOR
            \STATE $\tilde{\mathcal{B}} \gets \tilde{\mathcal{B}} \cup \{(\hat{x}_i, y_i)\}$
       \ENDFOR
         \STATE $\tilde{\mathcal{B}} \gets \tilde{\mathcal{B}} \cup \mathcal{B}$
        \STATE $\theta \leftarrow \theta - \frac{\eta}{2m}\cdot \nabla_{\theta} \sum_{(x, y) \in \tilde{\mathcal{B}}} \mathcal{L}_{\text{CE}}( y,f_\theta (x))$
        \ENDFOR
   \ENDFOR
   \STATE Output: Trained model $f_\theta$
\end{algorithmic}
\end{algorithm}

\subsection{Adversarial training with structured confidence perturbation (SCP)}

% To explicitly maximize the deformation of the confidence manifold, we propose SCP, which directly optimizes the distributional distance $\mathcal{D}$. We adopt the KL divergence as $\mathcal{D}$, which locally approximates the Mahalanobis distance on the manifold defined by the Fisher information matrix (FIM).

SCP transforms adversarial training into a geometric optimization problem on a confidence manifold.

% Assuming that the input of the model is $x \in \mathcal{X}$ and the parameter is $\theta$, the posterior probability distribution cluster output by the model constitutes a statistical manifold
% \begin{equation}
%     \small
%     \mathbb{P} = \{p(y|x;\theta)|x \in \mathcal{X}\},
% \end{equation}
% where each input $x$ corresponds to a point on the manifold, and the output space $\mathcal{X}$ constitutes a local coordinate system of the manifold.

% In adversarial training, the input sample $x$ is mapped to a point $p(y|x)$ on the manifold through the model, and the perturbation $x \to x + \delta$ will correspond to a trajectory on the manifold. Unlike the traditional method of maximizing the loss function, given the magnitude of the perturbation, we try to maximize the distance between two model output points on the manifold. We solve two sub-problems for iterative optimization:
% \begin{align}
%     \small
%     x'^* & = \arg\max_{x'} \mathcal{D}(p(y|x;\theta), p(y|x';\theta)), \label{eqn:maximization}\\
%     \theta^* & = \arg\min_{\theta} \mathbb{E}_{(x',y) \sim D} L(x',y;\theta), \label{eqn:minimization}
% \end{align}
% where $\mathcal{D}(\cdot,\cdot)$ represents the distance between the outputs of the two distribution functions.
% % , and unlike problem (\ref{eqn:optimal-delta}) and (\ref{eqn:min_loss}), the maximization problem has different objective functions from the minimization problem.

\subsubsection{Theoretical foundation} 

For the sake of convenience, we ignore the model parameter $\theta$. By the following theorem, we can conclude that the KL divergence between any two conditional distributions $p(y|x)$ and $p(y|x')$ is approximately equal to half the Mahalanobis distance between $x$ and $x'$, where the covariance parameter matrix is the inverse of the Fisher information matrix. The detailed derivation is provided in Appendix~\ref{app:theoretical-proofs}.

\begin{theorem}[Local approximation of KL divergence]
\label{thm:kl-divergence}
For any two class confidence distributions $p(y|x)$ and $p(y|x')$, where $x$ and $x'$ are the inputs of the model and $y$ is the class label of the model output, we have 
\begin{equation}
\mathrm{KL}(p(y|x),p(y|x')) \approx \frac{1}{2}(x' - x)^T F(x) (x' - x) =  \frac{1}{2}\delta^T F(x) \delta,
\end{equation}
where $F(x)$ is the Fisher information matrix (FIM) defined as follows
\begin{equation}
F(x) = \mathbb{E}_{p(y|x)}[\nabla_{x} \log p(y|x) \nabla_{x} \log p(y|x)^T].
\end{equation}
\end{theorem}

% FIM geometrically represents the curvature of the probability distribution manifold at the point $x$. From Theorem \ref{thm:kl-divergence}, we can easily see that the perturbation value $\delta$ in adversarial training is approximately equal to the principal eigenvector of the Fisher information matrix.

Theorem~\ref{thm:kl-divergence} reveals that the KL divergence between two nearby conditional distributions is locally approximated by the quadratic form $\frac{1}{2}\delta^\top F(x)\delta$. This establishes that the Fisher Information Matrix $F(x)$ defines a Riemannian metric on the statistical manifold of output distributions $\{p(y|x;\theta)\}$. The most sensitive perturbation direction at a given input $x$ is therefore the one that maximizes this KL divergence under a fixed Euclidean norm constraint $\|\delta\|_2 = \epsilon$:
\begin{equation}
    \max_{\|\delta\|_2 = \epsilon} \delta^\top F(x) \delta.
\end{equation}

By the Rayleigh--Ritz theorem, the maximum is attained when $\delta$ is aligned with the principal eigenvector of $F(x)$, and the maximal KL divergence equals $\frac{\epsilon^2}{2} \lambda_{\max}(F(x))$.

This geometric insight provides the foundation for SCP: rather than following the gradient of cross-entropy loss, which only suppresses the true class probability, we perturb along the direction that maximally deforms the entire confidence manifold. The FIM not only captures local curvature but also identifies the input direction that induces the most radical structural change in the output distribution, enabling a more principled approach to adversarial training.

\begin{theorem}[FIM for softmax model]
\label{thm:fisher-information}
For a deep learning model whose last layer uses a softmax function to implement classification tasks, where the input vector of softmax is $f(x)$, the Fisher information matrix is
\begin{equation}
    F(x)= \text{var}(\nabla_x f),
\end{equation}
where $\nabla_x f(x)$ is the gradient matrix (Jacobian matrix) of the vector $f(x)$ with respect to the input $x$ and $\text{var}$ represents the variance of the matrix random variable. The proof is provided in Appendix~\ref{app:fim-softmax-proof}.
\end{theorem}

%\begin{theorem}
%\texttt{(See Appendix Theorem 3)}
%\label{thm:eigen_decompose}

\subsubsection{Efficient algorithm design}
\label{sec:efficient-algorithm-design}
The Fisher information matrix of $x$ for the discrete variable $y$ is defined as
\begin{equation}\label{eqn:fx-decomposition}
    F(x) = \sum_{k = 1}^K p(y_k|x) [\nabla_{x} \log p(y_k|x) \nabla_{x} \log p(y_k|x)^T].
\end{equation}
Further, let $q_k = \nabla_{x} \log p(y_k|x)$  and $\lambda_k = p(y_k|x)$, we can write the Fisher information matrix in a more compressed form
\begin{equation}
    F(x) = Q \Lambda Q^T,
\end{equation}
where 
\begin{equation}
Q = \begin{bmatrix}
q_1 & q_2 & \cdots & q_K 
\end{bmatrix}
\in \mathbb{R}^{d \times K},
\end{equation}
and \begin{equation}
\Lambda = \begin{bmatrix}
\lambda_1 &  & &   \\
&  \lambda_2 &  &  \\
&  &  \ddots & \\ 
&  &     &  \lambda_K
\end{bmatrix}
\in \mathbb{R}^{K \times K}.
\end{equation}
Notice that $d$ is the dimension of the image $x$ and $K$ is the number of classes ($d \gg K$). It can be seen that the $(x' - x)$ (or $\delta$) that maximizes the KL divergence is the eigenvector of $F(x)$ associated with the largest eigenvalue.
However, directly storing or decomposing the Fisher matrix is very costly. 

Here we introduce a technique to solve this problem. While Fisher information matrix $F \in \mathbb{R}^{d \times d}$, we construct a matrix :
\begin{equation}
B = \Lambda^{1/2} Q^T Q \Lambda^{1/2}
\in \mathbb{R}^{K \times K}.
\end{equation}

% Let the eigenvector of $F$ be $v_F$, and the eigenvector of $B$ be $v_B$, the relation of $v_F$ and $v_B$ can be represented as follows: 
% \begin{equation}
% \small
% v_F = Q\Lambda^{1/2} v_B
% \label{eqn:va_to_vb} 
% \end{equation}

% Let $v_B$ be an eigenvector of $B$ with eigenvalue $\lambda$, i.e., $B v_B = \lambda v_B$. 
% Recall from Eqn. (\ref{eqn:fx-decomposition}) that $F = Q \Lambda Q^T$. Consider the vector $v_F = Q \Lambda^{1/2} v_B$. Then
% \begin{equation}
% F v_F = Q \Lambda Q^T (Q \Lambda^{1/2} v_B) = Q \Lambda (Q^T Q) \Lambda^{1/2} v_B.
% \end{equation}
% Since $B = \Lambda^{1/2} Q^T Q \Lambda^{1/2}$, we have $Q^T Q = \Lambda^{-1/2} B \Lambda^{-1/2}$. Substituting this into the expression yields
% \begin{equation}
% F v_F = Q \Lambda (\Lambda^{-1/2} B \Lambda^{-1/2}) \Lambda^{1/2} v_B = Q \Lambda^{1/2} B v_B = \lambda Q \Lambda^{1/2} v_B = \lambda v_F.
% \end{equation}
Let the eigenvector of $F$ be $v_F$,and the eigenvector of $B$ be $v_B$, the relation of $v_F$ and $v_B$ can be represented as follows: 
\begin{equation}
v_F = Q\Lambda^{1/2} v_B.
\label{eqn:va_to_vb} 
\end{equation}
% Thus $v_F$ is an eigenvector of $F$ with the same eigenvalue $\lambda$, and the relationship is
% \begin{equation}
% \label{eqn:va_to_vb}
% v_F = Q \Lambda^{1/2} v_B.
% \end{equation}

According to (\ref{eqn:va_to_vb}), we can simply decompose a $K\times K$ matrix $B$ instead of directly dealing with the Fisher Information Matrix. 

%\end{theorem}

To obtain the adversarial examples, we select the eigenvector corresponding to the largest eigenvalue, which is the most sensitive direction of the model. Finally, we get our adversarial algorithm SCP Alg. \ref{alg:scp}

\begin{algorithm}
\caption{Adversarial training with structured confidence perturbation (SCP)}
\label{alg:scp}
% \small
% \footnotesize

\begin{algorithmic}[1]
    % \scriptsize
   \STATE Input: DNN classifier $f_\theta$; dataset $\mathcal{D}$ with $K$ classes; batch size $m$; learning rate $\eta$; radius for adversaries $\epsilon$;number of epochs $T$
   \STATE Randomly initialize the network parameter $\theta$
   \FOR {$t = 0,1,2,\cdots,T$}
       \FOR {each batch $\mathcal{B} = \{(x_i, y_i)\}_{i=1}^m$ in $\mathcal{D}$}
        \STATE  $\tilde{\mathcal{B}} = \emptyset$
               \STATE $p(y_k|x_i; \theta) = \text{softmax}(f_\theta(x_i))$ 
               \STATE  $Q_i = [q_1, q_2, \ldots, q_K] $ , $q_k = \nabla_x \log p(y_k|x_i; \theta)$
               \STATE  $\Lambda_i = \text{diag}(p(y_1|x_i), p(y_2|x_i), \ldots, p(y_K|x_i))$

           \FOR {$(x_i, y_i)$ in $\mathcal{B}$}
               \STATE $B_i = \Lambda_i^{1/2} Q_i^T Q_i \Lambda_i^{1/2} $
               \STATE Decompose $B_i$ and find the principal eigenvector $v_{B_{i}}^*$ which corresponds to the largest eigenvalue $\lambda_{B_{i}}^*$
               \STATE $v_{F_i} = Q_i\Lambda_i^{1/2}v_{B_{i}}^*$
               \STATE $\hat{x}_i \leftarrow x_i + \epsilon \cdot\text{sign}(v_{F_i})$  
                 \STATE $\tilde{\mathcal{B}} \gets \tilde{\mathcal{B}} \cup \{(\hat{x}_i, y_i)\}$
           \ENDFOR
           
         \STATE $\tilde{\mathcal{B}} \gets \tilde{\mathcal{B}} \cup \mathcal{B}$
          \STATE $\mathcal{L}_{\text{SCP}} = \frac{1}{2m} \sum_{(x, y) \in \tilde{\mathcal{B}}} \mathcal{L}_{\text{CE}}( y,f_\theta(x))$
           \STATE $\theta \leftarrow \theta - \eta \nabla_\theta \mathcal{L}_{\text{SCP}}$
       \ENDFOR
   \ENDFOR
   \STATE Output: Trained model $f_\theta$
\end{algorithmic}
\end{algorithm}

Finally, we summarize the differences between our method and TRADES: TRADES combines clean cross-entropy with a KL robustness regularizer in the training objective, and its inner maximization generates perturbations by maximizing the KL divergence between clean and perturbed predictions. SPP instead directly maximizes CE + KL in the attack generation stage, actively disrupting distribution consistency, and then uses the generated examples for standard adversarial training. SCP further replaces the gradient direction geometrically with the principal eigenvector of the FIM.

  % \vspace{-20pt}
  \section{Experiments}

  \label{sec:Exp}
  \subsection{Datasets and settings} 
  \label{subsec:dataset}
  
  We used three common and popular datasets for image classification: CIFAR-10 \cite{krizhevsky2009learning}, CIFAR-100 \cite{krizhevsky2009learning}, and Tiny-ImageNet \cite{le2015tiny}. CIFAR-10 contains 60,000 color images of size 32 $\times$ 32 in 10 categories. CIFAR-100 contains 60,000 color images of size 32 $\times$ 32 in 100 categories. Tiny-ImageNet is a subset of ImageNet \cite{deng2009imagenet} with 200 categories, similar to the ImageNet dataset, where each category contains 500 color images of size 64 $\times$ 64.
  
  To comprehensively evaluate the relationship between model geometry and attack resistance, we conduct experiments on representative architectures with distinct structural characteristics:
  \begin{itemize}
  \item[$\bullet$] ResNet\cite{he2016deep}: It introduces residual connections, allowing gradients to bypass nonlinear layers directly. Skip connections stabilize the gradient flow, reduce gradient vanishing/explosion, and reduce variance, so its maximum eigenvalue is relatively low.
  \item[$\bullet$] DenseNet\cite{huang2017densely}: Each layer of DenseNet receives the output of all previous layers as input. With more gradient paths and gradients dispersed to each layer, its variance is potentially the smallest among the considered models.
  \item[$\bullet$] ViT\cite{dosovitskiy2020image}: Vision Transformer (ViT) is based on the self-attention mechanism, which causes the gradient to involve global interactions, potentially introducing large variance; however, LayerNorm and residual connections help stabilize the gradient.
  \end{itemize}
  
  The batch size on each dataset is 32, the learning rate $\eta$ is $3 \times 10^{-4}$
  , and the maximum perturbation range coefficient $\epsilon$ is $8/255$ in training section. The adversarial training contains 100 epochs. All experiments were conducted on a server equipped with six NVIDIA GeForce RTX 3090 GPUs. At the same time, we introduce CW\cite{carlini2017towards} and AutoAttack\cite{croce2020reliable} as our counter-test methods. We test our SCP method on multiple datasets and baseline models. The specific parameters of AutoAttack are $\epsilon= 0.5$, the type of norm is L2; the parameter settings of the CW-10 method are as follows: the learning rate is $0.01$, the confidence threshold is $0$, and the number of iterations is $10$.

  \subsection{Evaluation metrics}
  \label{subsec:metric}
  
  Assume that the test set is $D$, the set of all instances predicted correctly by the model $M$ is $T$, and $a(x)$ represents the perturbed sample generated by the clean input $x$. We can define the following three main metrics: STD, ADV, and ASR:
  \begin{itemize}
  \item[$\bullet$] Standard Accuracy (STD) \cite{lin2024hard} represents the standard test accuracy of in-distribution generalization, which is the accuracy obtained by testing the model using the original test sample, i.e, $\mathrm{STD} = \frac{|T|}{|D|}$.
  \item[$\bullet$] Adversarial Robust Accuracy (ADV) Adversarial robust accuracy is a method used to evaluate the performance of a model when facing adversarial samples. When we use a dataset consisting of perturbed samples generated by test samples to test the model obtained through adversarial training, the classification accuracy obtained is called ADV accuracy. In our experiments with attack methods AutoAttack, it is defined as $\mathrm{ADV} = \frac{|\sum_{(x,y) \in T} M(a(x)) = y|}{|D|}$.
  
  \item[$\bullet$] Attack Success Rate (ASR) ASR measures the proportion of samples that an attacker can cause the model to make incorrect predictions and is used to measure the robustness of the model. It is worth noting that for the above three measurements, using the ADV under AutoAttack, we have the following relationship $\mathrm{ASR} = 1 - \frac{\mathrm{ADV}}{\mathrm{STD}}$.
  % \item Adversarial Robust Accuracy (ADV) Adversarial robust accuracy is a method used to evaluate the performance of a model when facing adversarial samples. When we use a dataset consisting of perturbed samples generated by test samples to test the model obtained through adversarial training, the classification accuracy obtained is called ADV accuracy, i.e. $\mathrm{ADV} = \frac{|\sum_{(x,y) \in T} M(a(x)) = y|}{|D|}$.In our experiments, we evaluated the ADV under CW and AutoAttack. For both attack methods, we used the corresponding adversarial samples to evaluate the robustness of the model.
  
  % \item Attack Success Rate (ASR) ASR measures the proportion of samples that an attacker can cause the model to make incorrect predictions and is used to measure the robustness of the model. It is worth noting that for the above three measurements, we have the following relationship $\mathrm{ASR} = 1 - \frac{\mathrm{ADV}}{\mathrm{STD}}$.In our experiments, we use the ADV under AutoAttack to compute ASR.
  
  \end{itemize}

  % \subsection{Analysis of Model Robustness under SCP}
  
  % We evaluate the SCP geometric adversarial training paradigm on ResNet, DenseNet, VGGNet, and Vision Transformer (ViT) architectures on CIFAR10, MNIST, and Tiny-ImageNet.  Table \ref{tab:main-self} reveals the fundamental relationship between model geometry and attack resistance (see the Appendix for more results). Below we give some qualitative analysis based on the characteristics of the model and the conclusion of Theorem \ref{thm:fisher-information}:
  
  % \noindent VGG There is no skip connection in VGG, which relies on stacked convolutional layers and pooling layers. The gradient needs to be back-propagated through all layers, and the path is long, which is easy to accumulate variance. Therefore, the maximum eigenvalue of the gradient variance is high and the robustness is weak.
  
  % \noindent ResNet ResNet introduces residual connections, allowing gradients to bypass nonlinear layers directly. Skip connections stabilize the gradient flow, reduce gradient vanishing/explosion, and reduce variance, so its maximum eigenvalue is low.
  
  % \noindent DenseNet Each layer of DenseNet receives the output of all previous layers as input, with more gradient paths and gradients dispersed to each layer, which makes its variance possibly the smallest among the four models.
  
  % \noindent ViT ViT is based on the self-attention mechanism, which causes the gradient to involve global interactions, which may introduce large variance; but LayerNorm and residual connections stabilize the gradient.

  \subsection{Comparison on attack metrics}
  
  We compare our SCP algorithm with state-of-the-art (SOTA) adversarial training methods on three widely used image classification datasets (CIFAR-10, CIFAR-100 and Tiny-ImageNet) using ResNet18, ResNet34, and ResNet50 \cite{he2016deep} as baseline models. These architectures were chosen as the common intersection of models used in recent adversarial training literature, balancing computational efficiency and representativeness. Adversarial examples are generated with a perturbation budget ($\epsilon_{2}$) customized for each dataset and the performance is evaluated using STD, ADV, and ASR, where lower ASR indicates higher robustness.

\begin{table}[!htbp]
\centering
\scriptsize
\setlength{\tabcolsep}{1.2pt}
\renewcommand{\arraystretch}{1.12}
  \caption{Comparison of adversarial training methods on CIFAR-10, CIFAR-100, and Tiny-ImageNet. \textcolor{red}{Red}: best; \textcolor{blue}{blue}: second best.}
\label{tab:adversarial_results_modified}
   \resizebox{\linewidth}{!}{
\begin{tabular}{ll*{12}{c}}
  \toprule
  \multirow{2}{*}{\textbf{Dataset}} & \multirow{2}{*}{\textbf{Method}} &
  \multicolumn{4}{c}{\textbf{ResNet-18}} & \multicolumn{4}{c}{\textbf{ResNet-34}} & \multicolumn{4}{c}{\textbf{ResNet-50}} \\
  \cmidrule(lr){3-6} \cmidrule(lr){7-10} \cmidrule(lr){11-14}
   & &  STD $\uparrow$& CW-10 $\uparrow$& AA $\uparrow$& ASR $\downarrow$
   & STD $\uparrow$& CW-10$\uparrow$& AA$\uparrow$ &ASR $\downarrow$
   & STD$\uparrow$ & CW-10 $\uparrow$& AA$\uparrow$& ASR $\downarrow$\\
  \hline
  \multirow{10}{*}{\textbf{CIFAR-10}}
   & \textbf{PGD \cite{madry2019deeplearningmodelsresistant}}
   & 78.87\% & 46.20\% & 62.37\%&20.92\%
   & 77.74\% & 50.26\% & 64.24\%&17.37\%
   & 75.38\% & 48.30\% & 62.49\% &17.10\%\\
   & \textbf{TRADES \cite{zhang2019theoretically}}
   & 80.72\% & 48.74\% & 64.93\% &19.56\%
   & 82.47\% & 49.15\% & 67.31\% &18.38\%
   & 80.20\% & 50.09\% & 64.69\% &19.34\%\\
   & \textbf{MIFGSM \cite{dong2018boosting}}
   & 81.92\% & 50.98\% & 64.11\% &21.74\%
   & 81.35\% & 53.30\% & 65.06\% &20.02\%
   & 78.86\% & 49.39\% & 61.78\% &21.66\%
   \\
   & \textbf{NIFGSM \cite{lin2019nesterov}}
   & 81.28\% & 52.07\% & 65.08\% &19.93\%
   & \textcolor{blue}{82.63\%} & 52.26\% & 67.39\% &18.44\%
   & 81.33\% & 54.91\% & 67.13\% &17.46\%
   \\
   & \textbf{DIFGSM \cite{xie2019improving}}
    & 81.51\% & 50.18\% & 64.23\% &21.20\%
   & 80.76\% & 53.31\% & 67.03\% &17.00\%
   & 79.61\% & 52.47\% & 66.00\% &17.10\%
   \\
   & \textbf{JITTER \cite{schwinn2023exploring}}
    & 77.99\% & 50.02\% & 64.92\% &16.76\%
   & 80.07\% & 50.47\% & 65.79\% &17.83\%
   & 72.11\% & 47.03\% & 60.36\% &16.29\%
  \\
   & \textbf{PIFGSM\cite{GaoZhang2020PatchWise}}
   &  \textcolor{blue}{82.61\%} & 48.37\% & 64.22\% &22.26\%
   & 82.51\% & 52.29\% & 67.00\% &18.80\%
   & 80.40\% & 52.74\% &67.79\% &15.68\%
  \\
   & \textbf{VNIFGSM\cite{wang2021enhancing}}
   & 76.08\% & 47.53\% & 59.31\% &22.04\%
   & 74.50\% & 46.88\% & 59.27\%&20.44\%
   & 62.17\% & 41.34\% & 49.17\%&20.91\%
   \\
   \cmidrule{2-14}
   & \textbf{SPP(Ours)}
   & 81.90\% & \textcolor{red}{60.33\%} & \textcolor{blue}{70.15\%} &\textcolor{red}{14.35\%}
   & 82.49\% & \textcolor{red}{ 60.79\%} & \textcolor{blue}{70.66\%}& \textcolor{blue}{14.34\%}
   & \textcolor{blue}{83.12\%} & \textcolor{red}{61.10\%} & \textcolor{blue}{70.99\%} &\textcolor{blue}{14.59\%}
   \\
   & \textbf{SCP(Ours)} &
   \textcolor{red}{82.69\%} &  \textcolor{blue}{55.25\%} &  \textcolor{red}{70.47\%}  &\textcolor{blue}{14.78\%}
   &  \textcolor{red}{83.17\%} &  \textcolor{blue}{55.47\%} &  \textcolor{red}{71.50\%} &\textcolor{red}{14.03\%}
   &  \textcolor{red}{83.24\%} &  \textcolor{blue}{59.02\%} &  \textcolor{red}{72.50\%} &\textcolor{red}{12.90\%}\\
  \hline
  \multirow{10}{*}{\textbf{CIFAR-100}}
   & \textbf{PGD \cite{madry2019deeplearningmodelsresistant}}
   & 41.70\% & 21.57\% & 28.94\% &30.60\%
   & 40.58\% & 27.18\% & 28.41\%&29.99\%
   & 35.34\% & 25.57\% & 25.64\% &27.45\%
   \\
   & \textbf{TRADES \cite{zhang2019theoretically}}
   & 46.46\% & 28.98\% & 32.07\% &30.97\%
   & 44.50\% & 31.43\% & 32.66\% &\textcolor{red}{26.61\%}
   & 36.17\% & 22.21\% & 22.20\% &38.62\%
   \\
   & \textbf{MIFGSM \cite{dong2018boosting}}
   & \textcolor{blue}{48.47\%} & 22.09\% & 30.43\% &37.22\%
   & \textcolor{blue}{47.39\%} & 30.98\% & 31.85\% &32.79\%
   & 39.70\% & 27.93\% & 27.93\% &29.65\%
   \\
   & \textbf{NIFGSM \cite{lin2019nesterov}}
   & 46.19\% & 30.26\% & 31.93\% &30.87\%
   & 46.20\% & 30.26\% &31.29\% &32.27\%
   & 40.92\% & 25.18\% & 25.16\% &38.51\%
   \\
   & \textbf{DIFGSM \cite{xie2019improving}}
  & 47.42\% & 30.30\% & 31.89\% &32.75\%
   & 46.35\% &31.15\% & 32.28\% &30.36\%
   & 41.57\% & 28.77\% & 28.79\% &30.74\%
   \\
   & \textbf{JITTER \cite{schwinn2023exploring}}
   & 41.35\% & 25.66\% & 29.22\% &29.33\%
   & 42.76\% & 29.10\% &30.22\% &29.33\%
   & 32.45\% & 24.34\% & 24.39\% &\textcolor{blue}{24.84\%}
   \\
   & \textbf{PIFGSM\cite{GaoZhang2020PatchWise}}
   & \textcolor{red}{50.34\%} & 30.44\% & \textcolor{red}{32.66\%}&35.12\%
   & \textcolor{red}{49.10\%} & \textcolor{red}{32.54\%} & \textcolor{red}{33.68\%} &31.41\%
   & 40.83\% & 28.39\% & 28.45\% &30.32\%
   \\
   & \textbf{VNIFGSM\cite{wang2021enhancing}}
   & 35.22\% & 21.65\% & 24.16\% &31.40\%
   & 36.70\% & 23.80\% & 24.65\% &32.83\%
   & 31.25\% & 21.58\% & 21.56\% &31.01\%
  \\
    \cmidrule{2-14}
   & \textbf{SPP(Ours)}
   & 43.92\% & \textcolor{blue}{31.20\%} & 31.96\% &\textcolor{blue}{27.23\%}
   & 45.45\% & 30.51\% & 32.50\% &28.49\%
   & \textcolor{red}{42.76\%} & \textcolor{blue}{29.49\%} & \textcolor{blue}{30.70\%} &28.20\%
   \\
   & \textbf{SCP(Ours)}
   & 44.18\% & \textcolor{red}{32.01\%} & \textcolor{blue}{32.38\%} &\textcolor{red}{26.71\%}
   & 44.88\% & \textcolor{blue}{32.43\%} &\textcolor{blue}{32.90\%}&\textcolor{blue}{26.69\%}
   & \textcolor{blue}{42.25\%} & \textcolor{red}{31.96\%} & \textcolor{red}{31.86\%} &\textcolor{red}{24.59\%}\\
  \hline
  \multirow{10}{*}{\textbf{Tiny-ImageNet}}
   & \textbf{PGD \cite{madry2019deeplearningmodelsresistant}}
   & 30.55\% & 20.33\% & 20.32\% &33.49\%
   & 29.91\% & 21.95\% & 21.95\% &\textcolor{red}{26.61\%}
   & 28.87\% & 17.98\% & 17.97\% &37.76\%
   \\
   & \textbf{TRADES \cite{zhang2019theoretically}}
   & 33.29\% & 19.57\% & 19.57\% &41.21\%
   & 32.12\% & 19.89\% & 22.79\% &29.05\%
   & 33.09\% & 21.44\% & 21.44\% &35.21\%
   \\
   & \textbf{MIFGSM \cite{dong2018boosting}}
   & 34.06\% & 20.19\% & 20.73\% &39.14\%
   & \textcolor{red}{35.14\%} & 19.38\% & 21.11\% &39.93\%
   & 34.37\% & 20.97\% & 20.99\% &38.93\%
   \\
   & \textbf{NIFGSM \cite{lin2019nesterov}}
   & \textcolor{red}{36.43\%} & 19.50\% & 21.28\% &41.59\%
   & 32.29\% & 22.60\% & 22.59\% &30.04\%
   & 33.38\% & 20.26\% & 20.29\% &39.22\%
   \\
   & \textbf{DIFGSM \cite{xie2019improving}}
   & 34.70\% & 18.98\% & 22.83\% &34.21\%
   & 31.54\% & 22.72\% &22.71\% &28.00\%
   & \textcolor{blue}{34.49\%} & 22.86\% & 22.88\% &33.66\%
   \\
   & \textbf{JITTER \cite{schwinn2023exploring}}
   & 30.10\% & 21.25\% & 21.23\% &\textcolor{blue}{29.47\%}
   & 31.58\% & 22.95\% & 22.95\% &\textcolor{blue}{ 27.33\%}
   & 29.79\% & 21.08\% &21.08\% &29.24\%
   \\
   & \textbf{PIFGSM\cite{GaoZhang2020PatchWise}}
   & \textcolor{blue}{35.93\%} & 21.48\% & 22.09\% &38.52\%
   & 32.92\% & 23.07\% & 23.06\% &29.95\%
   & 33.82\% & \textcolor{blue}{24.55\%} & \textcolor{blue}{24.55\%} &\textcolor{red}{27.41\%}
   \\
   & \textbf{VNIFGSM\cite{wang2021enhancing}}
   & 25.66\% & 16.97\% & 16.99\% &33.79\%
   & 24.75\% & 17.00\% & 17.03\% &31.19\%
   & 20.56\% & 12.84\% & 12.83\% &37.60\%
  \\
    \cmidrule{2-14}
   &\textbf{SPP(Ours)}
   & 34.43\% & \textcolor{red}{24.13\%} & \textcolor{red}{24.30\%} &\textcolor{red}{29.42\%}
   & \textcolor{blue}{34.38\%} & \textcolor{red}{23.51\%} & \textcolor{red}{23.83\%} &30.69\%
   & \textcolor{red}{34.54\%} & \textcolor{red}{24.82\%} & \textcolor{red}{24.93\%} &\textcolor{blue}{27.82\%}
   \\
   & \textbf{SCP(Ours)}
   & 33.49\% & \textcolor{blue}{23.08\%} & \textcolor{blue}{23.20\%} &30.73\%
   & 33.33\% & \textcolor{blue}{23.32\%} & \textcolor{blue}{23.70\%} &28.89\%
   & 33.80\% & 24.02\% & 24.32\% &28.05\%
   \\
  \bottomrule
  \end{tabular}
  }
  \end{table}

  Driven by its geometric approach to perturbing the belief manifold rather than traditional scalar loss optimization, SCP demonstrates competitive performance, particularly on the CIFAR-10 and CIFAR-100 datasets. As shown in Table~\ref{tab:adversarial_results_modified}, SCP frequently achieves the highest robust accuracy under AutoAttack (AA). For instance, on CIFAR-10 with ResNet-50, SCP achieves an AA of $72.50\%$, surpassing SPP ($70.99\%$) and significantly outperforming baseline methods. Conversely, SPP exhibits superior performance on the Tiny-ImageNet dataset. Across all three ResNet architectures on Tiny-ImageNet, SPP yields higher Standard Accuracy (STD) than SCP. Its Attack Success Rate (ASR) is lower on ResNet-18 and ResNet-50; on ResNet-34, SPP has an ASR of $30.69\%$, compared with SCP's $28.89\%$.
  
  From a dataset perspective, SCP consistently minimizes the Attack Success Rate (ASR) compared to traditional methods like PGD and TRADES, validating its focus on structured belief perturbations over heuristic input space attacks. This advantage is most performed on CIFAR-10 and CIFAR-100. For example, using the ResNet-50 architecture on CIFAR-10, SCP achieves a remarkable ASR of $12.90\%$, which is a substantial improvement over TRADES ($19.34\%$) and PGD ($17.10\%$). Even on the more challenging CIFAR-100 dataset, SCP maintains this robust superiority; with ResNet-50, it suppresses the ASR to $24.59\%$, whereas competitive baselines like MIFGSM and PIFGSM struggle with ASRs of $29.65\%$ and $30.32\%$, respectively.
  
  The proposed methods, SPP and SCP, perform exceptionally well across varying model depths: ResNet-18, ResNet-34, and ResNet-50, demonstrating that the geometric disruption of the belief manifold scales effectively with network complexity. Unlike SOTA baselines such as PIFGSM, which witness a performance drop as model depth increases (e.g., dropping to $40.83\%$ STD on CIFAR-100/ResNet-50), the proposed methods exhibit enhanced adversarial resilience without suffering catastrophic drops in standard accuracy. On CIFAR-10/ResNet-50, both SPP and SCP maintain high clean accuracy (STD) of $83.12\%$ and $83.24\%$ respectively, while simultaneously securing the highest CW-10 and AA scores in the comparison. 
  
  On Tiny-ImageNet, SCP has lower STD than SPP at all three depths and higher ASR at two of the three depths. We attribute this behavior to three factors. First, the large number of classes (200) inflates the estimation variance of the Fisher Information Matrix, degrading the accuracy of its principal eigenvector. Second, the underfitting on this dataset may violate SCP's assumption of a well-converged confidence manifold, making its single-step perturbation unreliable. Third, SPP's multi-step signed gradient updates are inherently more robust to gradient noise and do not rely on explicit geometric estimation, offering greater stability under challenging training conditions.
  \subsection{Comparison with recent baselines}

  We further compare our methods with recent adversarial training baselines under the CIFAR-10 setting.

\begin{table}[!t]
  \centering
  \caption{Comparison with recent adversarial training baselines on CIFAR-10. \textcolor{red}{Red}: best; \textcolor{blue}{blue}: second best.}
  \label{tab:recent_baselines}
  \small
  \setlength{\tabcolsep}{3pt}
  \renewcommand{\arraystretch}{1.10}
  \begin{tabular}{lllccc}
  \toprule
  \textbf{Dataset} & \textbf{Method} & \textbf{Model} & STD$\uparrow$ & AA$\uparrow$ & ASR$\downarrow$ \\
  \midrule
 \multirow{8}{*}{\textbf{CIFAR-10}} & RPF \cite{dong2023adversarial} & PreActResNet-18 & 84.21\% & 53.89\% & 36.01\% \\
   & RAMP \cite{jiang2024rampboostingadversarialrobustness} & PreActResNet-18 & 82.10\% & 66.60\% & 18.88\% \\
   & GAWP \cite{jiang2024geometry} & PreActResNet-18 & \textcolor{red}{88.07\%} & 68.89\% & 21.78\% \\
   & LOAT
   \cite{yin2024boosting} & ResNet-18 & 77.97\% & 55.66\% & 28.61\% \\
   & RAT \cite{jin2023randomized} & ResNet-18 & 81.87\% & 60.19\% & 26.48\% \\
   & AR-AT \cite{waseda2025rethinking} & ResNet-18 & \textcolor{blue}{86.80\%} & 62.45\% & 28.05\% \\
  \cmidrule{2-6}
   & \textbf{SPP (Ours)} & ResNet-18 & 81.90\% & \textcolor{blue}{70.15\%} & \textcolor{red}{14.35\%} \\
   & \textbf{SCP (Ours)} & ResNet-18 & 82.69\% & \textcolor{red}{70.47\%} & \textcolor{blue}{14.78\%} \\
  \bottomrule
  \end{tabular}
  \renewcommand{\arraystretch}{1.0}
  \end{table}

 As shown in Table~\ref{tab:recent_baselines}, the proposed methods remain highly competitive against recent adversarial training baselines under the AutoAttack evaluation. SCP achieves the best robust accuracy of $70.47\%$, while SPP obtains the second-best result of $70.15\%$, both surpassing the strongest non-ours baseline GAWP ($68.89\%$). SPP and SCP also reduce ASR to $14.35\%$ and $14.78\%$, respectively, compared with RAMP's $18.88\%$. These results indicate that our confidence-geometry-based training improves robustness rather than merely increasing clean accuracy.

 % As shown in Table~\ref{tab:recent_baselines}, the proposed methods remain highly competitive against recent adversarial training baselines under the unified AutoAttack-L2 evaluation. SCP achieves the best AA-L2 accuracy of $70.47\%$, while SPP obtains the second-best result of $70.15\%$, both surpassing the strongest non-ours baseline GAWP-MART ($68.89\%$). This advantage is more evident under the same ResNet-18 setting: SCP improves AA-L2 accuracy over AR-AT/GAWP-MART ($62.45\%$), RAT-AWP ($60.19\%$), LOAT/LORE ($55.66\%$), and GAWP-AT ($52.76\%$) by $8.02$, $10.28$, $14.81$, and $17.71$ percentage points, respectively. Meanwhile, although GAWP-MART obtains the highest STD ($88.07\%$), its ASR remains $21.78\%$, and the best non-ours ASR is still $18.88\%$ from RAMP; in contrast, SPP and SCP reduce ASR to $14.35\%$ and $14.78\%$. These results indicate that our confidence-geometry-based training improves AA-L2 robustness rather than merely increasing clean accuracy.
   
  \subsection{Cross-architecture evaluation}
  
  To comprehensively evaluate the relationship between model geometry and attack resistance, we conduct experiments on representative architectures with distinct structural characteristics, as shown in Table~\ref{tab:architecture_comparison}:
  % \begin{itemize}
  % \item[$\bullet$]  ResNet \cite{he2016deep}: It introduces residual connections, allowing gradients to bypass nonlinear layers directly. Skip connections stabilize the gradient flow, reduce gradient vanishing/explosion, and reduce variance, so its maximum eigenvalue is relatively low.
  % \item[$\bullet$] DenseNet \cite{huang2017densely}: Each layer of DenseNet receives the output of all previous layers as input. With more gradient paths and gradients dispersed to each layer, its variance is potentially the smallest among the considered models.
  % \item[$\bullet$] ViT \cite{dosovitskiy2020image}: Vision Transformer (ViT) is based on the self-attention mechanism, which causes the gradient to involve global interactions, potentially introducing large variance; however, LayerNorm and residual connections help stabilize the gradient.
  % \end{itemize}

\begin{table}[!t]
  \centering
  \caption{Comparison of DenseNet121 and ViT-B/16 on CIFAR-10. \textcolor{red}{Red}: best; \textcolor{blue}{blue}: second best.}
  \label{tab:architecture_comparison}
  \small
  \setlength{\tabcolsep}{3pt}
  \renewcommand{\arraystretch}{1.10}
  \resizebox{\linewidth}{!}{
  \begin{tabular}{llcccccccc}
  \toprule
  \multirow{2}{*}{\textbf{Dataset}} & \multirow{2}{*}{\textbf{Method}} & \multicolumn{4}{c}{\textbf{DenseNet121}} & \multicolumn{4}{c}{\textbf{ViT-B/16}} \\
  \cmidrule(lr){3-6} \cmidrule(lr){7-10}
   & & STD$\uparrow$ & CW-10$\uparrow$ & AA$\uparrow$ & ASR$\downarrow$
   & STD$\uparrow$ & CW-10$\uparrow$ & AA$\uparrow$ & ASR$\downarrow$ \\
  \midrule
  \multirow{6}{*}{\textbf{CIFAR-10}}
   & TRADES \cite{zhang2019theoretically}
   & 76.24\% & 48.56\% & 62.27\% & 18.32\%
   & \textcolor{blue}{71.60\%} & 47.37\% & \textcolor{blue}{48.55\%} & 32.19\% \\
   & MIFGSM \cite{dong2018boosting}
   & 77.35\% & 49.79\% & 61.03\% & 21.10\%
   & \textcolor{red}{74.14\%} & \textcolor{red}{50.36\%} & \textcolor{red}{52.39\%} & \textcolor{blue}{29.34\%} \\
   & DIFGSM \cite{xie2019improving}
   & 77.75\% & 49.21\% & 62.68\% & 19.38\%
   & 70.90\% & \textcolor{blue}{47.43\%} & 48.53\% & 31.55\% \\
   & PIFGSM \cite{GaoZhang2020PatchWise}
   & 78.87\% & 49.41\% & 63.33\% & 19.70\%
   & 68.22\% & 45.13\% & 46.32\% & 32.10\% \\
  \cmidrule{2-10}
   & \textbf{SPP (Ours)}
   & \textcolor{blue}{81.62\%} & \textcolor{red}{60.38\%} & \textcolor{blue}{69.15\%} & \textcolor{red}{15.28\%}
   & 68.12\% & 44.55\% & 45.37\% & 33.40\% \\
   & \textbf{SCP (Ours)}
   & \textcolor{red}{82.49\%} & \textcolor{blue}{57.43\%} & \textcolor{red}{69.79\%} & \textcolor{blue}{15.40\%}
   & 66.20\% & 43.03\% & 46.94\% & \textcolor{red}{29.09\%} \\
  \bottomrule
  \end{tabular}
  }
  \renewcommand{\arraystretch}{1.0}
  \end{table}

  For DenseNet121, both SPP and SCP maintain their superiority, with SCP achieving an outstanding STD of 82.49\% and a robust AA of 69.79\%. This aligns with our expectations: DenseNet's dense skip connections disperse gradient flow across layers, effectively minimizing gradient variance and providing a stable, smooth geometric manifold that perfectly suits our covariance-based perturbation estimation.
  Conversely, on the ViT-B/16 architecture, we observe a systemic performance degradation across \textit{all} evaluated adversarial training methods, including our SPP/SCP and established baselines like TRADES and PIFGSM. The standard accuracy (STD) and robust accuracy (AA) for all methods drop significantly compared to their CNN counterparts. We attribute this phenomenon to two fundamental conflicts between the ViT architecture and current gradient-based adversarial training paradigms:
  \begin{itemize}
  \item[$\bullet$] Lack of Local Inductive Bias and Severe Underfitting: Unlike CNNs, ViTs lack translational invariance and local inductive biases. They are notoriously data-hungry and prone to severe underfitting when trained from scratch on small-scale datasets like CIFAR-10. Introducing strong adversarial constraints (whether through TRADES' KL-divergence or our geometric manifold perturbations) acts as an excessive regularization that further chokes the model's representation capacity, leading to the observed collapse in STD.
  \item[$\bullet$] High Gradient Variance and Manifold Instability: ViTs rely heavily on global self-attention mechanisms, meaning that spatial perturbations interact globally across the entire image patch sequence. This global interaction induces highly non-local and high-variance gradients. Since almost all current adversarial training methods (especially our SCP, which relies on the Fisher Information Matrix for local geometric approximation) depend on stable local gradient information, the noisy gradient landscape of ViT causes the estimated worst-case perturbation directions to become highly inaccurate. This instability explains why neither traditional PGD-based methods nor our covariance-guided framework can achieve optimal robustness on Vision Transformers without specialized architectural adaptations.
  \end{itemize}
  
  \subsection{Diagnostic analysis on ViT-B/16}
  
  To provide direct evidence for the above explanation, we further measure gradient variation on CIFAR-10. For each mini-batch $b$, we compute the mean input-gradient norm
  $g_{x}^{(b)}=\frac{1}{m}\sum_{i=1}^{m}\|\nabla_{x_i}\mathcal{L}_{b}\|_2$.
  Their relative variation is computed as
  \begin{equation}
  \mathrm{RV}(g)=\frac{\mathrm{std}_{b}(g^{(b)})}{\mathrm{mean}_{b}(g^{(b)})},
  \end{equation}
  where a larger value indicates a noisier gradient landscape across batches.
  
  % For FIM direction stability, we add a small random perturbation $\xi=10^{-3}$ to each test image and compare the principal FIM eigenvectors before and after the perturbation:
  % \begin{equation}
  % S(x)=|\cos(v_F(x),v_F(x+\xi))|.
  % \end{equation}
  % where $v_F(\cdot)$ denotes the principal FIM direction. A larger $S(x)$ means that the estimated local sensitive direction is more stable under small input perturbations.

  \begin{figure}[htbp]
      \centering
      \includegraphics[width=\linewidth]{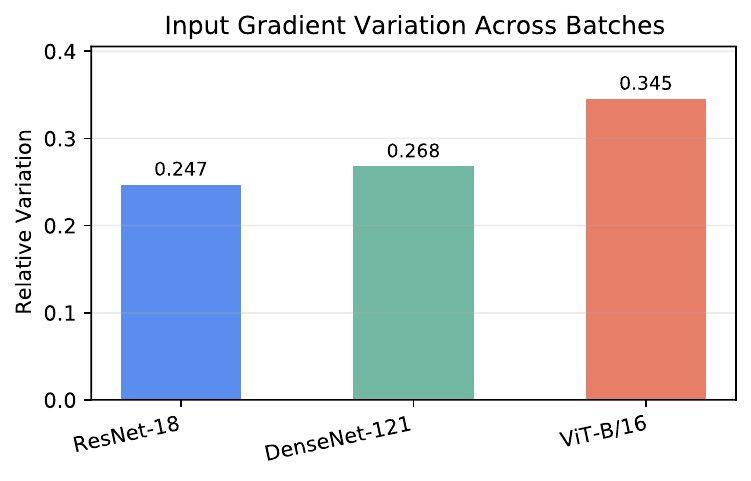}
      % \par\vspace{1pt}
      % {\footnotesizeGradient variation\par}
      % \vspace{8pt}
      % \includegraphics[width=0.75\linewidth]{fim_direction_stability.pdf}
      % \par\vspace{1pt}
      % {\footnotesize(b) FIM direction stability\par}
      \caption{Diagnostic analysis of gradient variation on CIFAR-10.}
      % \label{fig:vit_diagnostics}
  \end{figure}

  % \begin{figure}[htbp]
  % \centering
  % \begin{minipage}{0.49\linewidth}
  %     \centering
  %     \includegraphics[width=\linewidth]{gradient_variance.pdf}
  %     {\scriptsize(a)}
  % \end{minipage}
  % \hfill
  % \begin{minipage}{0.45\linewidth}
  %     \centering
  %     \includegraphics[width=\linewidth]{fim_direction_stability.pdf}
  %     {\scriptsize(b)}
  % \end{minipage}
  % \caption{Diagnostic analysis of ViT-B/16 degradation on CIFAR-10.}
  % \label{fig:vit_diagnostics}
  % \end{figure}    

  As shown in Fig.~2, ViT-B/16 exhibits the largest input-gradient variation ($0.345$), compared with ResNet-18 ($0.247$) and DenseNet-121 ($0.268$). It also shows a more dispersed FIM direction-stability distribution than the CNN architectures. These observations support our claim that ViT-B/16 produces noisier local gradients and less reliable local geometric estimates, which weakens SCP on this architecture.
  
  \subsection{Cross-model transferability}
  
  We next evaluate whether adversarial examples generated on one architecture can transfer to different target architectures. We use three ResNet-18 source models trained with PGD, TRADES, and PIFGSM, generate AutoAttack-L2 adversarial examples on CIFAR-10, and directly evaluate the same adversarial examples on DenseNet-121 and ViT-B/16 target models without re-optimizing them on the target models. This follows the standard black-box transfer setting \cite{papernot2016transferability}.
  
  Let $D$ be the test set, $M$ be a target model, and $x_{\mathrm{adv}}^{s,A}$ be the adversarial example generated by attack $A$ on a source model $s$. Since transfer attacks should only count samples that the target model originally classifies correctly, we first define the clean-correct subset
  \begin{equation}
  T_M=\{(x,y)\in D:M(x)=y\}.
  \end{equation}
  The transfer robust accuracy and attack success rate are computed as
  \begin{equation}
  \mathrm{Acc}_{s\rightarrow M}^{A}
  =
  \frac{|\{(x,y)\in T_M:M(x_{\mathrm{adv}}^{s,A})=y\}|}{|T_M|}.
  \end{equation}
Table~\ref{tab:resnet18_transfer_grouped} reports results for each combination of attack, source model, and target training method, where each entry is formatted as Robust Acc.
   % CW attack rows are temporarily commented out for compact presentation.
   % TRADES & 84.75/15.25 & 85.80/14.20 & 85.64/14.36 & 79.19/20.81 & 79.92/20.08 & 78.48/21.52 \\
   % MIFGSM & 83.28/16.72 & 84.04/15.96 & 82.65/17.35 & 79.52/20.48 & 80.49/19.51 & 79.12/20.88 \\
   % DIFGSM & 84.60/15.40 & 85.39/14.61 & 84.19/15.81 & 78.75/21.25 & 80.18/19.82 & 78.60/21.40 \\
   % PIFGSM & 84.48/15.52 & 84.77/15.23 & 83.97/16.03 & 78.44/21.56 & 80.03/19.97 & 78.79/21.21 \\
   % SPP (Ours) & 87.81/12.19 & 88.10/11.90 & 88.04/11.96 & 77.45/22.55 & 78.64/21.36 & 76.99/23.01 \\
   % SCP (Ours) & 88.07/11.93 & 88.19/11.81 & 88.57/11.43 & 81.96/18.04 & 82.55/17.45 & 81.90/18.10 \\
   % \midrule

\begin{table}[!t]
  \centering
  \caption{Cross-model transfer results on CIFAR-10. AutoAttack-L2 adversarial examples are generated on ResNet-18 source models and evaluated on DenseNet-121 and ViT-B/16 target models. Each entry reports robust accuracy $\uparrow$. The best and second-best results in each column are highlighted in \textcolor{red}{red} and \textcolor{blue}{blue}, respectively.}
  \label{tab:resnet18_transfer_grouped}
  \small
  \setlength{\tabcolsep}{3pt}
  \renewcommand{\arraystretch}{1.10}
  \begin{tabular}{lcccccc}
  \toprule
  \multirow{2}{*}{\textbf{Method}} &
  \multicolumn{3}{c}{\textbf{DenseNet121}} &
  \multicolumn{3}{c}{\textbf{ViT-B/16}} \\
  \cmidrule(lr){2-4} \cmidrule(lr){5-7}
   & \textbf{PGD} & \textbf{TRADES} & \textbf{PIFGSM}
     & \textbf{PGD} & \textbf{TRADES} & \textbf{PIFGSM} \\
  \midrule
   TRADES\cite{zhang2019theoretically}  & 92.94\% & 93.17\% & 92.63\% & 91.68\% & 92.29\%& 90.58\% \\
   MIFGSM \cite{dong2018boosting}& 92.41\%& 92.54\% & 90.59\%& \textcolor{blue}{92.03\%} & 92.58\% & 90.94\% \\
   DIFGSM \cite{xie2019improving} & 92.80\% & 92.99\% & 91.37\% & 92.02\% & \textcolor{blue}{92.87\%} & \textcolor{blue}{91.26\%} \\
   PIFGSM \cite{GaoZhang2020PatchWise} & 93.03\% & 92.61\%& 91.52\% & 91.83\% & 92.12\% & 90.73\% \\
   \midrule
   \textbf{SPP (Ours)} & \textcolor{blue}{94.23\%} & \textcolor{blue}{94.33\%} & \textcolor{blue}{93.89\%} & 91.13\% & 91.68\% & 90.22\% \\
   \textbf{SCP (Ours)} & \textcolor{red}{94.65\%} & \textcolor{red}{94.64\%} & \textcolor{red}{94.22\%} & \textcolor{red}{93.10\%} & \textcolor{red}{93.14\%} & \textcolor{red}{92.19\%} \\
  \bottomrule
  \end{tabular}
  \renewcommand{\arraystretch}{1.0}
  \end{table}

  As shown in Table~\ref{tab:resnet18_transfer_grouped}, target models trained with our methods are generally less affected by transferred adversarial examples. Under AutoAttack, SCP gives the best transfer robustness for both DenseNet-121 and ViT-B/16 across the reported source models, while SPP usually ranks second on DenseNet-121.
  
  \subsection{FIM spectrum analysis}
  
  To further inspect the geometric effect of different training methods, we compute the maximum eigenvalue $\lambda_{\max}(F)$ of the input Fisher Information Matrix on CIFAR-10. We use the low-dimensional matrix decomposition described in Sec.~\ref{sec:efficient-algorithm-design} to obtain the FIM spectrum efficiently. A smaller $\lambda_{\max}(F)$ means that the model's output distribution changes less sharply along its most sensitive local direction.

  % \begin{table}[!htbp]
  % \centering
  % \small
  % \caption{FIM maximum eigenvalue statistics on CIFAR-10 using ResNet-18.}
  % \label{tab:fim_spectrum_stats}
  % \scriptsize
  % \begin{tabular}{lccc}
  % \toprule
  % Method & Median $\lambda_{\max}\downarrow$ & $P(\lambda_{\max}\leq0.01)\uparrow$ & $P(\lambda_{\max}\leq0.1)\uparrow$ \\
  % \midrule
  % PGD & 0.1438 & 33.96\% & 47.62\% \\
  % TRADES & 0.3011 & 24.36\% & 40.93\% \\
  % PIFGSM & 0.1332 & 33.10\% & 47.84\% \\
  % MIFGSM & 0.2257 & 26.83\% & 43.39\% \\
  % SPP & 0.1131 & 32.61\% & 48.98\% \\
  % SCP & 0.0287 & 45.03\% & 56.06\% \\
  % \bottomrule
  % \end{tabular}
  % \end{table}

\begin{table}[!htbp]
\centering
\caption{FIM maximum eigenvalue statistics on CIFAR-10 using ResNet-18.}
\label{tab:fim_spectrum_stats}
\begin{tabular}{lccc}
\toprule
Method & Median $\lambda_{\max}\downarrow$ & $P(\lambda_{\max}\leq0.01)\uparrow$ & $P(\lambda_{\max}\leq0.1)\uparrow$ \\
\midrule
PGD & 0.1438 & 33.96\% & 47.62\% \\
TRADES & 0.3011 & 24.36\% & 40.93\% \\
PIFGSM & 0.1332 & 33.10\% & 47.84\% \\
MIFGSM & 0.2257 & 26.83\% & 43.39\% \\
SPP & 0.1131 & 32.61\% & 48.98\% \\
SCP & \textbf{0.0287} & \textbf{45.03\%} & \textbf{56.06\%} \\
\bottomrule
\end{tabular}
\end{table}

  As shown in Table~\ref{tab:fim_spectrum_stats}, SCP produces the largest fraction of samples with low curvature. This supports the geometric intuition that SCP training flattens the most sensitive direction of the confidence manifold, reducing sharp local confidence changes.

  \subsection{Visualization of the confidence shift}
  
  \label{subsec:visualization}
\begin{figure}[htbp]
      \centering
      % \vspace{-5pt}
      % 使用 0.5\linewidth 并配合 % 符号消除中间空格，实现无缝并排且最大化
      \begin{minipage}{0.48\linewidth}  
          \centering    
      \includegraphics[width=\linewidth,keepaspectratio]{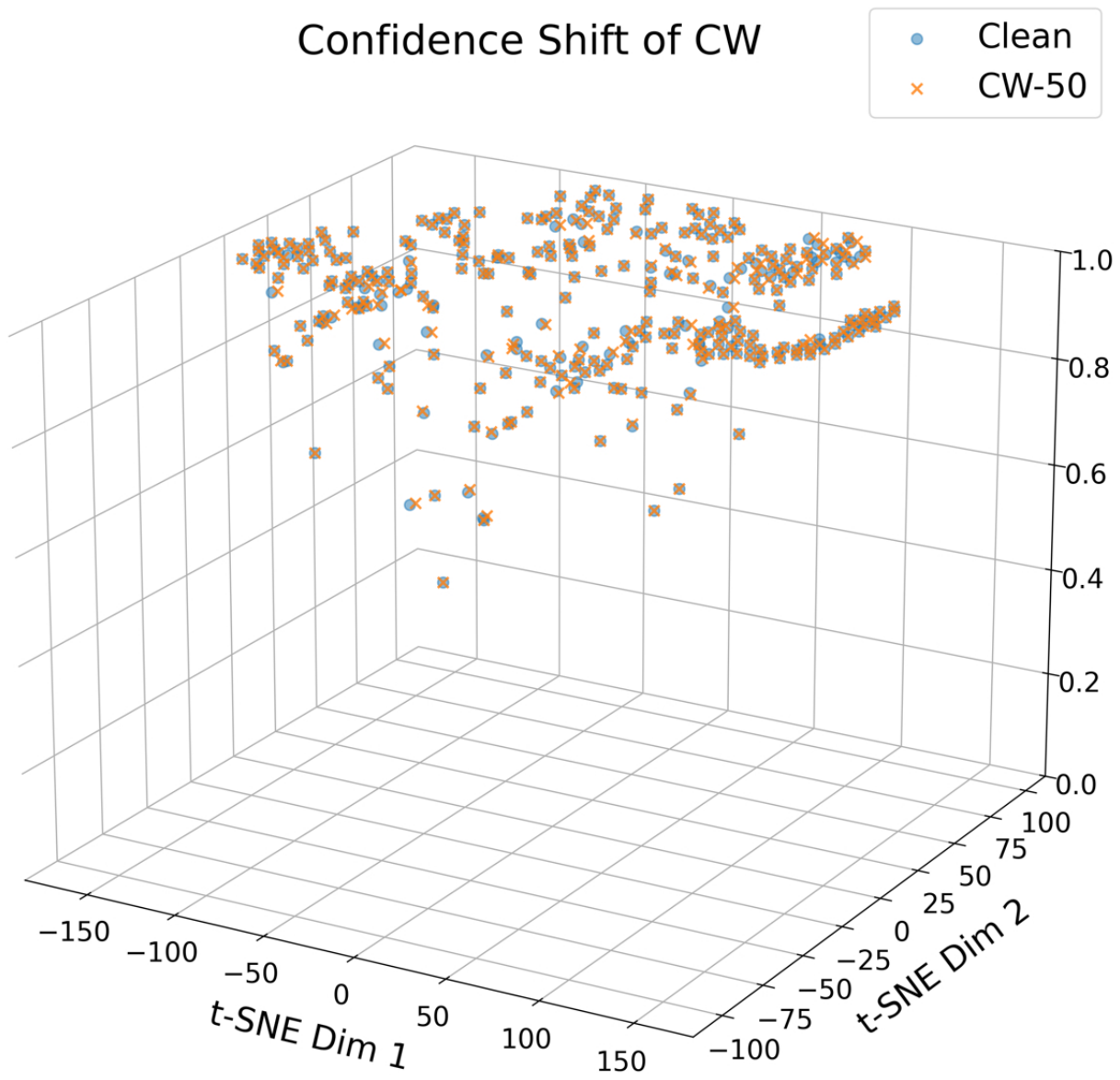}%
      \end{minipage}%
      \begin{minipage}{0.48\linewidth}
          \centering
          \includegraphics[width=\linewidth,keepaspectratio]{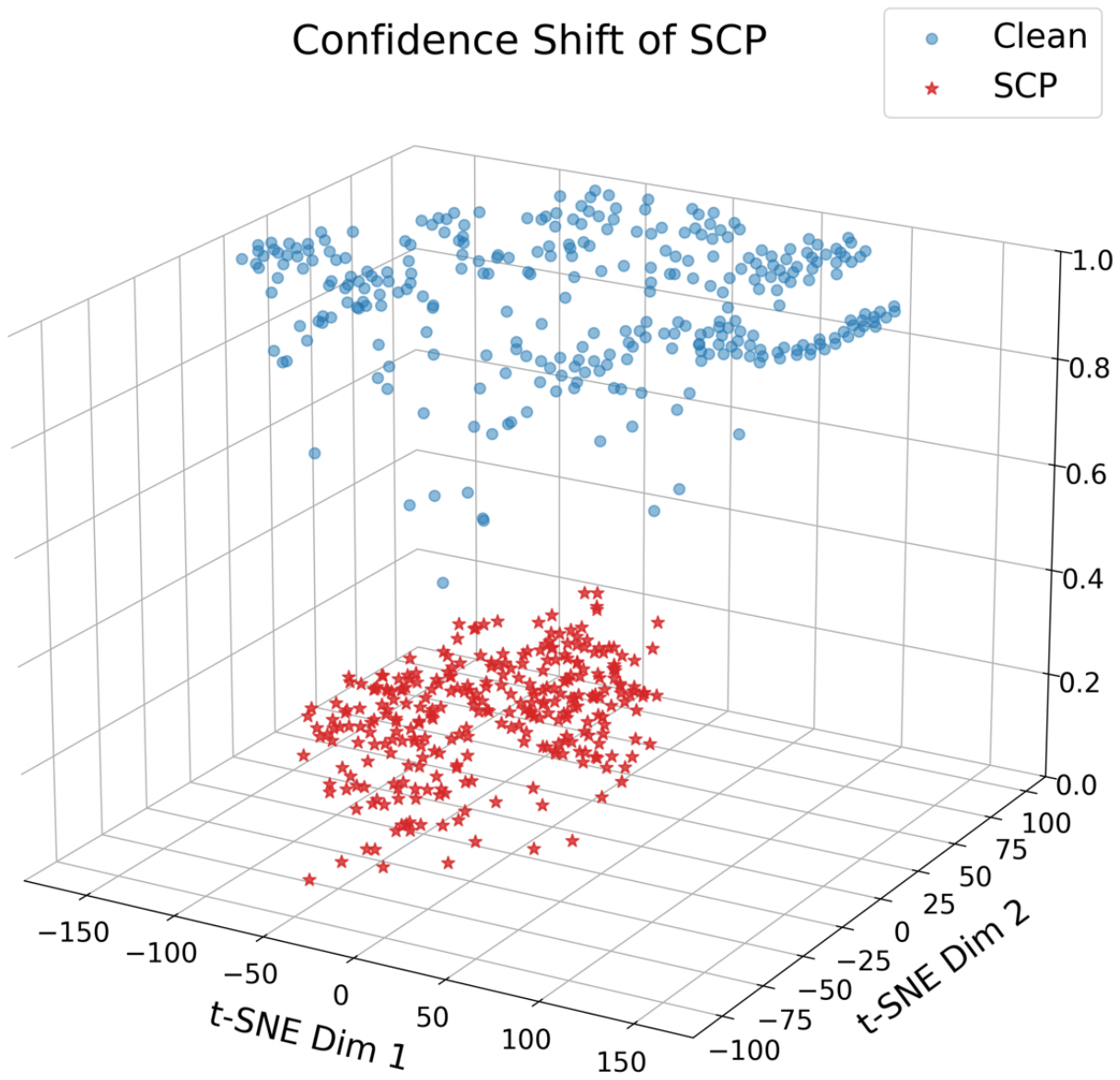}%
      \end{minipage}
      % \vspace{-3pt} 
      \caption{3D visualization comparison of confidence shift on CIFAR-10 (left: CW, right: SCP). The $X$-$Y$ plane represents t-SNE projected features, and the $Z$-axis represents the prediction confidence.}
      % \vspace{-10pt}
      \label{fig:cluster}
  \end{figure}

  To validate the primary motivation of our Structured Confidence Perturbation (SCP) framework---specifically,we utilize t-SNE \cite{JMLR:v9:vandermaaten08a} to project the high-dimensional latent features from the penultimate layer of a ResNet-18 model into a 2D coordinate system ($X$ and $Y$ axes). The $Z$-axis represents the model's prediction confidence, defined as the maximum softmax probability of the predicted class. Figure \ref{fig:cluster} illustrates the distribution shift between clean samples and adversarial examples generated by CW\cite{carlini2017towards} and our SCP method on the CIFAR-10 dataset.
  
  As shown in Figure \ref{fig:cluster}, benign samples (represented by blue points) cluster into distinct regions and reside on a high-confidence plateau, with $Z$-axis values consistently near 1.0. CW attack lead to many adversarial points remaining in high-confidence zones (represented by orange points). This phenomenon confirms the existence of \textit{pseudo-attacks}, where the perturbation fails to disrupt the model's belief structure fundamentally. In contrast, our SCP method induces a much more radical deformation of the confidence landscape (represented by red points). The confidence scores for SCP-generated samples drop sharply, indicating that our method effectively explores the most vulnerable regions of the statistical manifold.
  % \vspace{-5pt}
  % \begin{figure}[htbp]
  %     \centering
  %     \includegraphics[width=1.0\linewidth]{cluster_confidence_3d.png}
  %     \vspace{-20pt}
  %     \caption{3D visualization comparison of confidence shift on CIFAR-10. The $X$-$Y$ plane represents t-SNE projected features, and the $Z$-axis represents the prediction confidence.}
  %     \vspace{-15pt}
  %     \label{fig:cluster}
  % \end{figure}
  % \begin{figure}[htbp]
  %     \centering
     
  %     % 使用 0.5\linewidth 并配合 % 符号消除中间空格，实现无缝并排且最大化
  %     \begin{minipage}{0.55\linewidth}  
  %         \centering    
  %     \includegraphics[width=1\linewidth, ]{cw.pdf}%
  %     \end{minipage}%
  %     \begin{minipage}{0.55\linewidth}
  %         \centering
  %         \includegraphics[width=1\linewidth, ]{scp.pdf}%
  %     \end{minipage}
  
  %     \caption{3D visualization comparison of confidence shift on CIFAR-10 (left: CW, right: SCP). The $X$-$Y$ plane represents t-SNE projected features, and the $Z$-axis represents the prediction confidence.}
  
  %     \label{fig:cluster}
  % \end{figure}

  \subsection{Time and space complexity}
  We analyze the computational efficiency of SPP and SCP to demonstrate their practical feasibility. While SPP follows a standard linear complexity, the primary bottleneck for SCP has been solved by using the decomposition technique; see Eq.~\eqref{eqn:va_to_vb}.

  \begin{table}[!htbp]
  \centering
\small
  \caption{Time and space complexity of SPP and SCP with direct and indirect decomposition.}
  \begin{tabular}{lcc}
  \toprule
  Method & Space complexity & Time complexity \\
  \midrule
  SPP & $O(d)$ & $O(d)$ \\
  SCP (direct)&  $O(d^2)$ & $O(Kd^2 + d^3)$ \\
  SCP (indirect) & $O(dK)$ & $O(dK^2 + K^3)$ \\
  \bottomrule
  \end{tabular}
  \label{tab:time_complexity}
  \end{table}

  As shown in Table~\ref {tab:time_complexity}, our optimization reduces SCP's complexity, since $K\ll d$ (K=10/100/200, d=3072/12288) in most classification tasks. This approach ensures that our geometric paradigm remains scalable and feasible for large-scale deep learning systems.
  
  \subsection{Ablation study}
  
  Effectiveness of Hybrid loss: 
  To verify the effectiveness of the proposed hybrid loss function, we present a comparison in Table~\ref {tab:ablation_spp_loss} among the PGD (CE only), the TRADES (KL only), and our proposed hybrid loss SPP (CE+KL) on the CIFAR10 dataset using ResNet-18.
  
  As shown in Table~\ref{tab:adversarial_results_modified}, the proposed hybrid loss strategy achieves superior performance across all metrics compared with using either loss component individually. The SPP (CE+KL) method achieves the highest Clean Accuracy of 81.90\% and the highest AutoAttack accuracy of 70.15\%. This represents a significant improvement over the CE-only baseline and also outperforms the KL-only approach. In terms of ASR, the hybrid loss achieves the strongest defense, reducing ASR to 14.35\%. This is notably lower than both the CE-only (20.92\%) and KL-only (19.56\%) configurations.
  
  These results confirm that combining CE and KL losses allows the model to learn more robust features without compromising standard accuracy, demonstrating the effectiveness of the SPP hybrid loss design.

  \begin{table}[!htbp]
    \centering
\small
    \caption{Comparison of the hybrid loss SPP(CE+KL) with PGD(CE) and TRADES(KL) on CIFAR10 using ResNet-18.}
    \label{tab:ablation_spp_loss}
    \begin{tabular}{lccc}
      \toprule
      Loss Type      & Clean Acc$\uparrow$ & AA  $\uparrow$& ASR $\downarrow$\\
      \midrule
      CE only        &  78.87\%        &  62.37\%      & 20.92\%   \\
      KL only        &  80.72\%        &  64.93\%      & 19.56\%   \\
      CE + KL (SPP)  &  \textbf{81.90\%}& \textbf{70.15\%} & \textbf{14.35\%} \\
      \bottomrule
        \end{tabular}
  \end{table}

  We further analyze the sensitivity of SPP to the KL-loss weight. In this experiment, the SPP training objective is
  \begin{equation}
  \mathcal{L}_{\mathrm{SPP}}=\mathcal{L}_{\mathrm{CE}}+\lambda\mathcal{L}_{\mathrm{KL}},
  \end{equation}
  where $\lambda$ controls the contribution of the distribution-consistency term. We train ResNet-18 on CIFAR-10 with $\lambda \in \{0.1,0.5,1.0,2.0\}$ and evaluate the trained models under clean, PGD, CW-10, and AutoAttack-L2 settings.

  \begin{figure}[htbp]
      \centering
      \includegraphics[width=\linewidth]{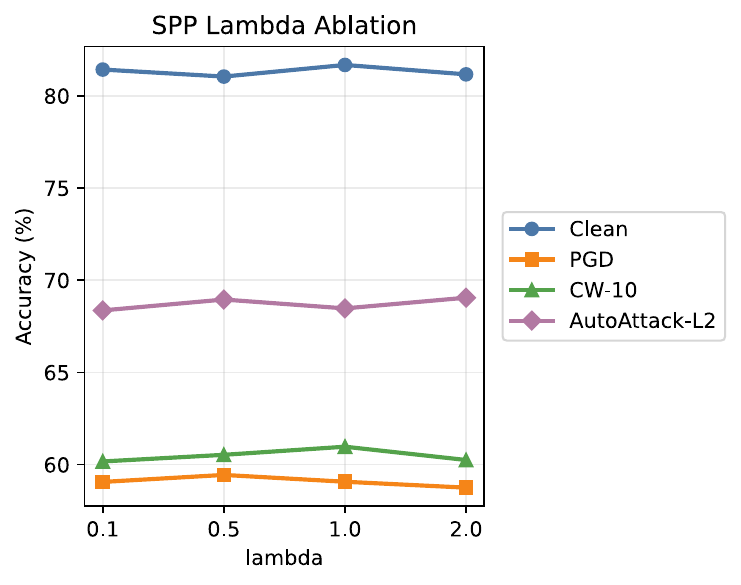}
      \caption{SPP lambda ablation on CIFAR-10 using ResNet-18.}
      \label{fig:spp_lambda_ablation}
  \end{figure}

  As shown in Fig.~\ref{fig:spp_lambda_ablation}, the SPP performance is stable across a broad range of $\lambda$. The clean and attack accuracies fluctuate only mildly, indicating that the CE and KL terms are complementary rather than requiring fragile tuning. In particular, $\lambda=1.0$ gives the best clean and CW-10 performance, while $\lambda=2.0$ achieves the best accuracy.

\FloatBarrier
\section{Conclusion}
\label{sec:conclu}\label{sec:conclusion}

We address the ``pseudo-attack'' problem using two complementary approaches---Single Point Perturbation (SPP) and Structured Confidence Perturbation (SCP)---systematically attacking and strengthening the model's confidence structure stability from two levels: local sample-level perturbation and global manifold-level deformation, respectively. SPP integrates a hybrid loss function combining CE and KL, while SCP further models adversarial perturbation as a distance maximization problem on the confidence manifold from an information geometry perspective. SCP's scalable computation makes it suitable for large-scale classification tasks. Experimental results show that the proposed method significantly improves adversarial robustness on multiple standard datasets and model architectures. Visualization analysis further confirms that SCP effectively induces global deformation of the confidence distribution, preventing adversarial examples from remaining in high-confidence regions. Future research directions include exploring its potential to improve robustness in generative models and self-supervised learning.

% The PR guide requires numbered references in order of first citation.
% The CAS bundle's author-year BST sorts alphabetically; use Elsevier's numeric BST.

\label{main-body-end}
\clearpage
\bibliographystyle{unsrtnat}
\bibliography{icdm2026references}

@inproceedings{wang2020improving,
title={Improving Adversarial Robustness Requires Revisiting Misclassified Examples},
author={Yisen Wang and Difan Zou and Jinfeng Yi and James Bailey and Xingjun Ma and Quanquan Gu},
booktitle={ICLR},
year={2020},
}

@article{li2022review,
author = {Yao Li, Minhao Cheng, Cho-Jui Hsieh and Thomas C. M. Lee},
title = {A Review of Adversarial Attack and Defense for Classification Methods},
journal = {The American Statistician},
year = {2022},
}

@article{huang2017adversarial,
  title={Adversarial attacks on neural network policies},
  author={Huang, Sandy and Papernot, Nicolas and Goodfellow, Ian and Duan, Yan and Abbeel, Pieter},
  journal={arXiv},
  year={2017}
}

@article{chakraborty2021survey,
author = {Chakraborty, Anirban and Alam, Manaar and Dey, Vishal and Chattopadhyay, Anupam and Mukhopadhyay, Debdeep},
title = {A survey on adversarial attacks and defences},
year = {2021},
journal = {CAAI Transactions on Intelligence Technology},
}

@article{kabid2023autonomous,
  title={Towards autonomous driving model resistant to adversarial attack},
  author={Shibly, Kabid Hassan and Hossain, Md Delwar and Inoue, Hiroyuki and Taenaka, Yuzo and Kadobayashi, Youki},
  journal={Applied Artificial Intelligence},
  year={2023}
}

@article{zhu2022adversarial,
title = {Adversarial training of LSTM-ED based anomaly detection for complex time-series in cyber-physical-social systems},
journal = {Pattern Recognition Letters},
year = {2022},
author = {Haiqi Zhu and Shaohui Liu and Feng Jiang},
}

@inproceedings{he2016deep,
  title={Deep residual learning for image recognition},
  author={He, Kaiming and Zhang, Xiangyu and Ren, Shaoqing and Sun, Jian},
  booktitle={CVPR},
  year={2016}
}

@inproceedings{huang2017densely,
  title={Densely connected convolutional networks},
  author={Huang, Gao and Liu, Zhuang and Van Der Maaten, Laurens and Weinberger, Kilian Q},
  booktitle={CVPR},
  year={2017}
}

@article{dosovitskiy2020image,
  title={An image is worth 16x16 words: Transformers for image recognition at scale},
  author={Dosovitskiy, Alexey and Beyer, Lucas and Kolesnikov, Alexander and Weissenborn, Dirk and Zhai, Xiaohua and Unterthiner, Thomas and Dehghani, Mostafa and Minderer, Matthias and Heigold, Georg and Gelly, Sylvain and others},
  journal={arXiv:2010.11929},
  year={2020}
}

@techreport{krizhevsky2009learning,
  title={Learning multiple layers of features from tiny images},
  author={Krizhevsky, Alex},
  year={2009},
  institution={Citeseer}
}

@inproceedings{papernot2016transferability,
  title={Transferability in Machine Learning: from Phenomena to Black-Box Attacks using Adversarial Samples},
  author={Papernot, Nicolas and McDaniel, Patrick and Goodfellow, Ian and Jha, Somesh and Celik, Z. Berkay and Swami, Ananthram},
  booktitle={arXiv:1605.07277},
  year={2016}
}

@article{goodfellow2014explaining,
  title={Explaining and harnessing adversarial examples},
  author={Goodfellow, Ian J and Shlens, Jonathon and Szegedy, Christian},
  journal={arXiv:1412.6572},
  year={2014}
}

@misc{madry2019deeplearningmodelsresistant,
      title={Towards Deep Learning Models Resistant to Adversarial Attacks}, 
      author={Aleksander Madry and Aleksandar Makelov and Ludwig Schmidt and Dimitris Tsipras and Adrian Vladu},
      year={2019},
      archivePrefix={arXiv}
}

@article{shafahi2019adversarial,
  title={Adversarial training for free!},
  author={Shafahi, Ali and Najibi, Mahyar and Ghiasi, Mohammad Amin and Xu, Zheng and Dickerson, John and Studer, Christoph and Davis, Larry S and Taylor, Gavin and Goldstein, Tom},
  journal={NeurIPS},
  year={2019}
}

@article{zhang2019you,
  title={You only propagate once: Accelerating adversarial training via maximal principle},
  author={Zhang, Dinghuai and Zhang, Tianyuan and Lu, Yiping and Zhu, Zhanxing and Dong, Bin},
  journal={NeurIPS},
  year={2019}
}

@article{tramer2017ensemble,
  title={Ensemble adversarial training: Attacks and defenses},
  author={Tram{\`e}r, Florian and Kurakin, Alexey and Papernot, Nicolas and Goodfellow, Ian and Boneh, Dan and McDaniel, Patrick},
  journal={arXiv:1705.07204},
  year={2017}
}

@article{zhang2019defense,
  title={Defense against adversarial attacks using feature scattering-based adversarial training},
  author={Zhang, Haichao and Wang, Jianyu},
  journal={NeurIPS},
  year={2019}
}

@article{le2015tiny,
  title={Tiny imagenet visual recognition challenge},
  author={Le, Ya and Yang, Xuan},
  journal={CS 231N},
  year={2015}
}

@inproceedings{deng2009imagenet,
  title={Imagenet: A large-scale hierarchical image database},
  author={Deng, Jia and Dong, Wei and Socher, Richard and Li, Li-Jia and Li, Kai and Fei-Fei, Li},
  booktitle={CVPR},
  year={2009}
}

@inproceedings{zhang2019theoretically,
  title={Theoretically principled trade-off between robustness and accuracy},
  author={Zhang, Hongyang and Yu, Yaodong and Jiao, Jiantao and Xing, Eric and El Ghaoui, Laurent and Jordan, Michael},
  booktitle={ICML},
  year={2019}
}

@article{liu1989limited,
  title={On the limited memory BFGS method for large scale optimization},
  author={Liu, Dong C and Nocedal, Jorge},
  journal={Mathematical programming},
  year={1989}
}

@article{duchi2015optimal,
  title={Optimal rates for zero-order convex optimization: The power of two function evaluations},
  author={Duchi, John C and Jordan, Michael I and Wainwright, Martin J and Wibisono, Andre},
  journal={IEEE Transactions on Information Theory},
  year={2015}
}

@article{kraidia2024defense,
  title={Defense against adversarial attacks: robust and efficient compressed optimized neural networks},
  author={Kraidia, Insaf and Ghenai, Afifa and Belhaouari, Samir Brahim},
  journal={Scientific Reports},
  year={2024}
}

@article{zhou2022adversarial,
  title={Adversarial attacks and defenses in deep learning: From a perspective of cybersecurity},
  author={Zhou, Shuai and Liu, Chi and Ye, Dayong and Zhu, Tianqing and Zhou, Wanlei and Yu, Philip S},
  journal={ACM Computing Surveys},
  year={2022}
}

@inproceedings{carlini2017towards,
  title={Towards Evaluating the Robustness of Neural Networks},
  author={Carlini, Nicholas and Wagner, David},
  booktitle={IEEE SP},
  year={2017},
}

@inproceedings{jin2023randomized,
  title={Randomized adversarial training via taylor expansion},
  author={Jin, Gaojie and Yi, Xinping and Wu, Dengyu and Mu, Ronghui and Huang, Xiaowei},
  booktitle={CVPR},
  year={2023}
}

@inproceedings{dong2023adversarial,
  title={Adversarial robustness via random projection filters},
  author={Dong, Minjing and Xu, Chang},
  booktitle={CVPR},
  year={2023}
}

@misc{jiang2024rampboostingadversarialrobustness,
      title={RAMP: Boosting Adversarial Robustness Against Multiple $l_p$ Perturbations for Universal Robustness}, 
      author={Enyi Jiang and Gagandeep Singh},
      year={2024},
      eprint={2402.06827},
      archivePrefix={arXiv},
}

@article{jiang2024geometry,
  title={Geometry-aware weight perturbation for adversarial training},
  author={Jiang, Yixuan and Chiang, Hsiao-Dong},
  journal={Electronics},
  year={2024},
  publisher={MDPI}
}

@inproceedings{yin2024boosting,
  title={Boosting adversarial training via fisher-rao norm-based regularization},
  author={Yin, Xiangyu and Ruan, Wenjie},
  booktitle={CVPR},
  year={2024}
}

@inproceedings{waseda2025rethinking,
  title={Rethinking invariance regularization in adversarial training to improve robustness-accuracy trade-off},
  author={Waseda, Futa and Chang, Ching-Chun and Echizen, Isao},
  booktitle={ICLR},
  year={2025}
}

@article{andriushchenko2020understanding,
  title={Understanding and improving fast adversarial training},
  author={Andriushchenko, Maksym and Flammarion, Nicolas},
  journal={NeurIPS},
  year={2020}
}

@inproceedings{wu2025rethinking,
  title={Rethinking kullback-leibler divergence in knowledge distillation for large language models},
  author={Wu, Taiqiang and Tao, Chaofan and Wang, Jiahao and Yang, Runming and Zhao, Zhe and Wong, Ngai},
  booktitle={ACL},
  year={2025}
}

@article{lin2024hard,
  title={Hard Adversarial Example Mining for Improving Robust Fairness},
  author={Lin, Chenhao and Ji, Xiang and Yang, Yulong and Li, Qian and Zhao, Zhengyu and Peng, Zhe and Wang, Run and Fang, Liming and Shen, Chao},
  journal={IEEE TIFS},
  year={2024}
}

@inproceedings{croce2020reliable,
    title = {Reliable evaluation of adversarial robustness with an ensemble of diverse parameter-free attacks},
    author = {Francesco Croce and Matthias Hein},
    booktitle = {ICML},
    year = {2020}
}

@inproceedings{dong2018boosting,
  title={Boosting adversarial attacks with momentum},
  author={Dong, Yinpeng and Liao, Fangzhou and Pang, Tianyu and Su, Hang and Zhu, Jun and Hu, Xiaolin and Li, Jianguo},
  booktitle={CVPR},
  year={2018}
}

@inproceedings{GaoZhang2020PatchWise,
  author    = {Lianli Gao and
               Qilong Zhang and
               Jingkuan Song and
               Xianglong Liu and
               Heng Tao Shen},
  title     = {Patch-Wise Attack for Fooling Deep Neural Network},
  Booktitle = {ECCV},
  year      = {2020}
}

@inproceedings{wang2021enhancing,
  title={Enhancing the transferability of adversarial attacks through variance tuning},
  author={Wang, Xiaosen and He, Kun},
  booktitle={CVPR},
  year={2021}
}

@article{lin2019nesterov,
  title={Nesterov accelerated gradient and scale invariance for adversarial attacks},
  author={Lin, Jiadong and Song, Chuanbiao and He, Kun and Wang, Liwei and Hopcroft, John E},
  journal={arXiv:1908.06281},
  year={2019}
}

@inproceedings{xie2019improving,
  title={Improving transferability of adversarial examples with input diversity},
  author={Xie, Cihang and Zhang, Zhishuai and Zhou, Yuyin and Bai, Song and Wang, Jianyu and Ren, Zhou and Yuille, Alan L},
  booktitle={CVPR},
  year={2019}
}

@article{schwinn2023exploring,
  title={Exploring misclassifications of robust neural networks to enhance adversarial attacks},
  author={Schwinn, Leo and Raab, Ren{\'e} and Nguyen, An and Zanca, Dario and Eskofier, Bjoern},
  journal={Applied intelligence},
  year={2023},
  publisher={Springer}
}

@article{JMLR:v9:vandermaaten08a,
  author  = {Laurens van der Maaten and Geoffrey Hinton},
  title   = {Visualizing Data using t-SNE},
  journal = {JMLR},
  year    = {2008},
}

\clearpage
\appendix
\numberwithin{equation}{section}
\numberwithin{figure}{section}
\numberwithin{table}{section}
\setcounter{theorem}{0}
\renewcommand{\thetheorem}{\Alph{section}.\arabic{theorem}}
\section{Theoretical proofs}
\label{app:theoretical-proofs}

\subsection{Theoretical proofs on KL divergence under general discrete distribution}

\begin{theorem}
\phantomsection
\label{proof:kl-divergence}

For any two discrete probability distributions $p(y|x)$ and $p(y|x')$, we have 

\begin{equation}
\mathrm{KL}(p(y|x),p(y|x')) \approx \frac{1}{2}(x' - x)^T F(x) (x' - x),
\end{equation}

where

\begin{equation}
F(x) = \mathbb{E}_{p(y|x)}[\nabla_{x} \log p(y|x) \nabla_{x} \log p(y|x)^T].
\end{equation}

\begin{proof}
First, we perform Taylor's second-order expansion of the function $\log p(y|x')$ at point $x$

\begin{align}
\log p(y|x') &\approx \log p(y|x) + (\nabla_x \log p(y|x))^T (x' - x) \nonumber \\
&\quad + \frac{1}{2}(x' - x)^T \nabla_x^2 \log p(y|x)(x' - x) \label{eqn:taylor-expansion},
\end{align}

then substitute it into the KL divergence 

\begin{equation}
\mathbb{D}(p(y|x)||p(y|x'))=\sum_{i = 1}^K p(y_i|x)[\log p(y_i|x) - \log p(y_i|x')] \nonumber
\end{equation}

to get

\begin{align}
&\mathbb{D}(p(y|x)||p(y|x'))  =  -(x' - x)^T \sum_{i = 1}^K p(y_i|x) \nabla \log p(y_i|x) \\
&\quad - \frac{1}{2}(x' - x)^T \bigg(\sum_{i = 1}^K p(y_i|x) \nabla^2 \log p(y_i|x) \bigg) \nonumber (x' - x). \nonumber
\end{align}

For the first term above, we have

\begin{equation}
\sum_{i = 1}^K p(y_i|x) \nabla \log p(y_i|x) = \nabla \sum_{i = 1}^K p(y_i|x) = 0. \nonumber
\end{equation}

For the second term above, we have

\begin{align}
\nabla_x \log p(y_i|x) &= \frac{\nabla_x p(y_i|x)}{p(y_i|x)}, \nonumber \\
\nabla^2_x \log p(y_i|x) &= \frac{p(y_i|x)\nabla^2_x p(y_i|x)  - \nabla_x p(y_i|x) \nabla_x p(y_i|x)^T }{p(y_i|x)^2} \nonumber \\
&= \frac{\nabla^2_x p(y_i|x)}{p(y_i|x)} \nonumber  - \nabla_x \log p(y_i|x) \nabla_x \log p(y_i|x)^T, \nonumber 
\end{align}

further we get

\begin{align}
&\sum_{i = 1}^K p(y_i|x) \nabla^2_x \log p(y_i|x)\\
&= \sum_{i = 1}^K \Big\{\nabla^2_x p(y_i|x) - p(y_i|x)[\nabla_x \log p(y_i|x) \nabla_x \log p(y_i|x)^T] \Big\}, \nonumber \\
&= \nabla^2_x \sum_{i = 1}^K p(y_i|x) \nonumber- \mathbb{E}_{p(y|x)}[\nabla_x \log p(y|x) \nabla_x \log p(y|x)^T], \nonumber \\
&= - \mathbb{E}_{p(y|x)}[\nabla_x \log p(y|x) \nabla_x \log p(y|x)^T] = - F. \nonumber
\end{align}

Finally, we arrive at our conclusion.
\end{proof}

\end{theorem}

\subsection{Proof of the Fisher information matrix for deep learning models}
\label{app:fim-softmax-proof}
\begin{theorem}
\phantomsection
\label{proof:fim-softmax}
\begin{proof}

According to Theorem 1, the Fisher information matrix $F$ measures the sensitivity of the model output distribution $p(y|x)$ to the input $x$. For classification tasks, $F$ is defined as

\begin{equation}
F = \mathbb{E}_{p(y|x)}[\nabla_{x} \log p(y|x) \nabla_{x} \log p(y|x)^T].
\end{equation}

Next we need to estimate the maximum eigenvalue of the Fisher information matrix for some models.

For classification, we assume the model outputs $K$ probabilities $y_i$:
\begin{equation}\label{eq:softmax}
y_i = p(y = i|x) = \frac{e^{f_i(x)}}{\sum_{k = 1}^K e^{f_k(x)}},
\end{equation}
then
\begin{equation}\label{eq:log-softmax}
\log p(y = i|x) = f_i(x) - \log \sum_{k = 1}^K e^{f_k(x)}.
\end{equation}
Its gradient with respect to the input $x$ is (let $f_i = f_i(x)$):
\begin{align}\label{eq:gradient-log-softmax}
\nabla_x \log p(y = i|x) &= \nabla_x f_i - \sum_{k = 1}^K \left(\sum_{j = 1}^K e^{f_j} \right)^{-1} e^{f_k}\nabla_x f_k \nonumber \\
&= \nabla_x f_i - \sum_{k = 1}^K p(y = k|x)\nabla_x f_k \nonumber \\
&= \nabla_x f_i - \mathbb{E}_{p(y|x)}[\nabla_x f].
\end{align}
The covariance formula is defined as:
\begin{align}\label{eq:covariance}
\text{Cov}(X)& = \mathbb{E}[(X - \mathbb{E}[X])(X - \mathbb{E}[X])^T]. \\
 &=\mathbb{E}[(XX^T])-  \mathbb{E}[X]\mathbb{E}[X]^T]
\end{align}
Let $\mu = \mathbb{E}_{p(y|x)}[\nabla_x f]$ denote the mean gradient, then the Fisher information matrix can be represented as follows :
\begin{align}\label{eq:fisher-info}
F &= \sum_{k = 1}^K p(y = k|x)[(\nabla_x f_k - \mu)(\nabla_x f_k - \mu)^T] \nonumber \\
&= \mathbb{E}_{p(y|x)}[(\nabla_x f - \mu)(\nabla_x f - \mu)^T] \nonumber \\
&= \mathbb{E}[\nabla_x f\nabla_x f^T] - \mathbb{E}[\nabla_x f]\mathbb{E}[\nabla_x f]^T \nonumber \\
& = \text{var}(\nabla_x f)
\end{align}
\end{proof}
\end{theorem}

\begin{theorem}
\phantomsection
\label{proof:eigen-decomposition}
\begin{proof}
The Fisher information matrix of $x$ for the discrete variable $y$ is defined as
\begin{equation}
F(x) = \sum_{k = 1}^K p(y_k|x) [\nabla_{x} \log p(y_k|x) \nabla_{x} \log p(y_k|x)^T].
\end{equation}
Further, let $q_k = \nabla_{x} \log p(y_k|x)$  and $\lambda_k = p(y_k|x)$, we can write the Fisher information matrix in a more compressed form
\begin{equation}
F(x) = Q \Lambda Q^T,
\end{equation}
where 
\begin{equation}
Q = \begin{bmatrix}
q_1 & q_2 & \cdots & q_K 
\end{bmatrix}
\in \mathbb{R}^{d \times K}
\end{equation}
and \begin{equation}
\Lambda = \begin{bmatrix}
\lambda_1 &  & &   \\
&  \lambda_2 &  &  \\
&  &  \ddots & \\ 
&  &     &  \lambda_K
\end{bmatrix}
\in \mathbb{R}^{K \times K}
\end{equation}
Where $d$ is the dimension of the image $x$ and $K$ is the number of the classes($d \gg K$).While Fisher information matrix $F \in \mathbb{R}^{d \times d} $,we construct a matrix :
\begin{equation}
B = \Lambda^{1/2} Q^T Q \Lambda^{1/2}
\in \mathbb{R}^{K \times K}
\end{equation}

\begin{align}
    B v_B & = \lambda_{B} v_B \\
    \Lambda^{1/2} Q^T Q \Lambda^{1/2}v_B &= \lambda_{B} v_B \\ 
    Q\Lambda Q^T  (Q\Lambda^{1/2}v_B) &= \lambda_{B} Q\Lambda^{1/2} v_B\\
    F (Q\Lambda^{1/2}v_B) &=  \lambda_{B} (Q\Lambda^{1/2} v_B)
\end{align}
Then we have:
\begin{equation}
 v_F = Q\Lambda^{1/2} v_B
\end{equation}
\end{proof}
\end{theorem}
\normalsize
% \end{thebibliography}

\end{document}